\documentclass[fleqn,10pt]{wlpeerj} 

\newcommand{\erD}{\textit{\textbf{Multi-Emotion Regression Network}}  (ER)\xspace} 
\newcommand{\erS}{ER\xspace} 

\newcommand{\ecD}{\textit{\textbf{Single-Emotion Classification Network}}  (EC)\xspace} 
\newcommand{\ecS}{EC\xspace} 

\newcommand{\ecbpD}{\textit{\textbf{Binarized} Emotion-Vector \textbf{Behavior Primitives}} (B-BP)\xspace} 
\newcommand{\ecbpS}{B-BP\xspace} 

\newcommand{\erbpD}{\textit{Emotion-\textbf{Embedding Behavior Primitives} (E-BP)\xspace}} 
\newcommand{\erbpS}{E-BP\xspace} 

\newcommand{\bpD}{\emph{\textbf{Basic Affect Behavioral Primitive Information}} (or \emph{Behavioral Primitives} for short, BP)\xspace} 
\newcommand{\bpM}{behavioral primitives\xspace} 
\newcommand{\bpS}{BP\xspace} 

\usepackage{multirow}
\usepackage{tabularx}
\usepackage{amsmath}
\usepackage{graphicx}
\usepackage{comment}
\graphicspath{ {./img/} }
\usepackage{subfig}
\usepackage{soul}

\usepackage[utf8]{inputenc} 
\usepackage[T1]{fontenc}    
\usepackage{color}
\definecolor{linkcolor}{cmyk}{0.838,0.720,0.059,0.055}
\usepackage[colorlinks=true, linkcolor=linkcolor,
anchorcolor=linkcolor,
citecolor=linkcolor,
filecolor=linkcolor,
menucolor=linkcolor,
runcolor=linkcolor]{hyperref}       
\usepackage{url}            
\usepackage{booktabs}       
\usepackage{amsfonts}       
\usepackage{nicefrac}       
\usepackage{microtype}      
\usepackage{multirow}
\usepackage{nameref}
\usepackage[user,titleref]{zref}

\definecolor{mypink1}{rgb}{0.858, 0.188, 0.478}
\newcommand{\hq}[1]{{\color{mypink1}{\emph{#1}}}}
\renewcommand{\hq}[1]{}

\definecolor{myblue1}{rgb}{0.257, 0.5, 0.95}
\newcommand{\cm}[1]{\emph{\color{myblue1}#1:\ }}
\renewcommand{\cm}[1]{}

\usepackage{xspace}

\newcommand{\namerefQ}[1]{`{\ztitleref{#1}}'} 
\newcommand{\eg}{\textit{e.g.,\ }}
\newcommand{\etc}{\textit{etc.\@}}

\newcommand{\ie}{\textit{i.e.,\ }}

\newcommand{\panos}[1]{{\color{orange}#1}}
\usepackage{comment}
\specialcomment{panosc}{\color{red}}{\color{black}}

\title{Linking emotions to behaviors through deep transfer learning}

\author[1]{Haoqi Li}
\author[2]{Brian Baucom}
\author[1]{Panayiotis Georgiou}

\affil[1]{Department of Electrical and Computer Engineering, University of Southern California, Los Angeles, California, United States}
\affil[2]{Department of Psychology, University of Utah, Salt Lake City, Utah, United States}
\corrauthor[1]{Haoqi Li}{haoqili@usc.edu}

\begin{abstract}
Human behavior refers to the way humans act and interact.
Understanding human behavior is a cornerstone of observational practice, especially in psychotherapy. An important cue of behavior analysis is the dynamical changes of emotions during the conversation. Domain experts integrate emotional information in a highly nonlinear manner, thus, it is challenging to explicitly quantify the relationship between emotions and behaviors. 
In this work, we employ deep transfer learning to analyze their inferential capacity and contextual importance.
We first train a network to quantify emotions from acoustic signals and then use information from the emotion recognition network as features for behavior recognition. We treat this emotion-related information as behavioral primitives and further train higher level layers towards behavior quantification. Through our analysis, we find that emotion-related information is an important cue for behavior recognition. Further, we investigate the importance of emotional-context in the expression of behavior by constraining (or not) the neural networks' contextual view of the data. This demonstrates that the sequence of emotions is critical in behavior expression. To achieve these frameworks we employ hybrid architectures of convolutional networks and recurrent networks to extract emotion-related behavior primitives and facilitate automatic behavior recognition from speech.
\end{abstract}

\begin{document}
\flushbottom
\maketitle
\thispagestyle{empty}

\section*{Introduction}\zlabel{sec:intro}

Human communication includes a range of cues from lexical, acoustic and prosodic, turn taking and emotions to complex behaviors. 
Behaviors encode many domain-specific aspects of the internal user state, from highly complex interaction dynamics to expressed emotions.
These are encoded at multiple resolutions, time scales, and with different levels of complexity.
For example, a short speech signal or a single uttered word can convey basic emotions \citep{ekman1992_are-there-basic,ekman1992_an-argument-for}.
More complex behaviors require domain specific knowledge and longer observation windows for recognition. 
This is especially true in task specific behaviors of interest in observational treatment for psychotherapy such as in couples’ therapy \citep{christensen2004traditional} and suicide risk assessment \citep{cummins2015review}. 
Behaviors encompass a rich set of information that includes the dynamics of interlocutors  and their emotional states, and can often be domain specific. 
The evaluation and identification of domain specific behaviors (e.g. blame, suicide ideation) can facilitate effective and specific treatments by psychologists.
During the observational treatment, annotation of human behavior is a time consuming and complex task. 
Thus, there have been efforts on automatically recognizing human emotion and behavior states, which resulted in vibrant research topics such as affective computing \citep{tao2005affective, picard2003affective, sander2014oxford}, social signal processing \citep{vinciarelli2009social}, and behavioral signal processing (BSP) \citep{narayanan2013behavioral, georgiou2011behavioral}. 
In the task of speech emotion recognition (SER), researchers are combining machine learning techniques to build reliable and accurate affect recognition systems \citep{schuller2018speech}.
In the BSP domain, through domain-specific focus on areas such as human communication, mental health and psychology, research targets advances of understanding of higher complexity constructs and helps psychologists to observe and evaluate domain-specific behaviors.

However, despite these efforts on automatic emotion and behavior recognition (see \namerefQ{sec:related_work}), there has been less work on examining the relationship between these two. In fact, many domain specific annotation manuals and instruments \citep{heavey2002couples, jones1998couples, heyman2004rapid} have clear descriptions that state specific basic emotions can be indicators of certain behaviors. 
Such descriptions are also congruent with how humans process information. 
For example, when domain experts attempt to quantify complex behaviors, they often employ affective information within the context of the interaction at varying timescales to estimate behaviors of interest \citep{narayanan2013behavioral, tseng2016_couples-behavio}.

Moreover, the relationship between behavior and emotion provides an opportunity for (i) transfer learning by employing emotion data, that is easier to obtain, annotate, and less subjective, as the initial modeling task; and (ii) employing emotional information as building blocks, or primitive features, that can describe behavior. 

The purpose of this work is to explore the relationship between emotion and behavior through deep neural networks, and further the employ emotion-related information towards behavior quantification.
There are many notions of what an ``emotion'' is. For the purpose of this paper and most research in the field \citep{el2011survey,schuller2018speech}, the focus is on \emph{basic emotions}, which are defined as cross-culturally recognizable. One commonly used discrete categorization is by \citet{ekman1992_are-there-basic,ekman1992_an-argument-for}, in which six basic emotions are identified as anger, disgust, fear, happiness, sadness, and surprise. 
According to theories \citep{WorthPsychology, scherer2005emotions}, emotions are states of feeling that result in physical and psychological changes that influence our behaviors.

Behavior, on the other hand, encodes many more layers of complexity: the dynamics of the interlocutors, their perception, appraisal, and expression of emotion, their thinking and problem-solving intents, skills and creativity, the context and knowledge of interlocutors, and their abilities towards emotion regulation \citep{baumeister2007emotion, baumeister2010does}.
Behaviors are also domain dependent. In addiction \citep{baer2009agency}, for example, a therapist will mostly be interested in the language which reflects changes of addictive habits. 
In suicide prevention \citep{cummins2015review}, reasons for living and emotional bond are more relevant. In doctor-patient interactions, empathy or bedside manners are more applicable. 

In this paper, we will first address the task of basic emotion recognition from speech. Thus we will discuss literature on the notion of emotion (see \namerefQ{sec:def_emo}) and prior work on emotion recognition (see \namerefQ{sec:work_emo}). We will then, as our first scientific contribution, describe a system that can label emotional speech (see \namerefQ{subsec:emotionRecognition}). 

The focus of this paper however is to address the more complex task of behavior analysis. Given behavior is very related to the dynamics, perception, and expression of emotions \citep{WorthPsychology}, we believe a study is overdue in establishing the degree to which emotions can predict behavior.
We will therefore introduce more analytically the notion of behavior (see \namerefQ{sec:def_beh}) and describe prior work in behavior recognition (see \namerefQ{sec:work_beh}), mainly from speech signals.
The second task of this paper will be in establishing a model that can predict behaviors from basic emotions. 
We will investigate the emotion-to-behavior aspects in two ways: we will first assume that the discrete emotional labels directly affect behavior (see \namerefQ{subsubsec:behRecognition_seq}). 
We will further investigate if an embedding from the emotion system, representing behaviors but encompassing a wider range of information, can better encode behaviorally meaningful information (see. \namerefQ{subsubsec:e2b-embed}).

In addition, the notion that behavior is highly dependent on emotional expression also raises the question of how important the sequence of emotional content is in defining behavior. 
We will investigate this through progressively removing the context from the sequence of emotions in the emotion-to-behavior system (see. \namerefQ{subsec:e2b-reduced_context}) and study how this affects the automatic behavior classification performance.

\section*{Background} \zlabel{sec:background}
\paragraph{Emotions:}\zlabel{sec:def_emo}
\cm{What is Emotion}
There is no consensus in the literature on a specific definition of emotion.
An ``emotion'' is often taken for granted in itself and, most often, is defined with reference to a list of descriptors such as anger, disgust, happiness, and sadness \etc \citep{cabanac2002emotion}. 
\citet{oatley1996understanding} distinguish emotion from mood or preference by the duration of each kind of state.
Two emotion representation models are commonly employed in practice \citep{schuller2018speech}. 
One is based on the discrete emotion theory, 
where six basic emotions are isolated from each other, and researchers assume that any emotion can be represented as a mixture of the basic emotions \citep{cowie2001emotion}.
The other model defines emotions via continuous values corresponding to different dimensions which assumes emotions change in a continuous manner and have strong internal connections but blurred boundaries between each other. 
The two most common dimensions are arousal and valence \citep{schlosberg1954three}.

In our work, following related literature, we will refer to basic emotions as emotions that are expressed and perceived through a short observation window. Annotations of such emotions take place without context to ensure that time-scales, back-and-forth interaction dynamics, and domain-specificity is not captured.

\paragraph{Behavior:}\zlabel{sec:def_beh}
\cm{What is Behavior}
Behavior is the output of information and signals including but not limited to those: (i) manifested in both overt and covert multimodal cues (``expressions''); and (ii) processed and used by humans explicitly or implicitly (``experience'' and ``judgment'') \citep{narayanan2013behavioral, baumeister2010does}. 
Behaviors encompass significant degrees of emotional perception, facilitation, thinking, understanding and regulation, and are functions of dynamic interactions \citep{baumeister2007emotion}. Further, such complex behaviors are increasingly domain specific and subjective.

\paragraph{Link between emotions and behavior:}\label{sec:link_beh_emo}
Emotions can change frequently and quickly in a short time period \citep{ekman1992_are-there-basic, mower2011hierarchical}. 
They are internal states that we perceive or express (\eg through voice or gesture) but are not interactive and actionable. 
Behaviors, on the other hand, include highly complex dynamics, information from explicit and implicit aspects, are exhibited over longer time scales, and are highly domain specific. 

For instance, ``happiness'', as one of the emotional states, is brought about by generally positive feelings. While within couples therapy domain, behavior ``positivity'' is defined in \citep{heavey2002couples,jones1998couples} as ``Overtly expresses warmth, support, acceptance, affection, positive negotiation''.
\begin{figure}[ht]
    \centering
    \includegraphics[width=0.95\linewidth]{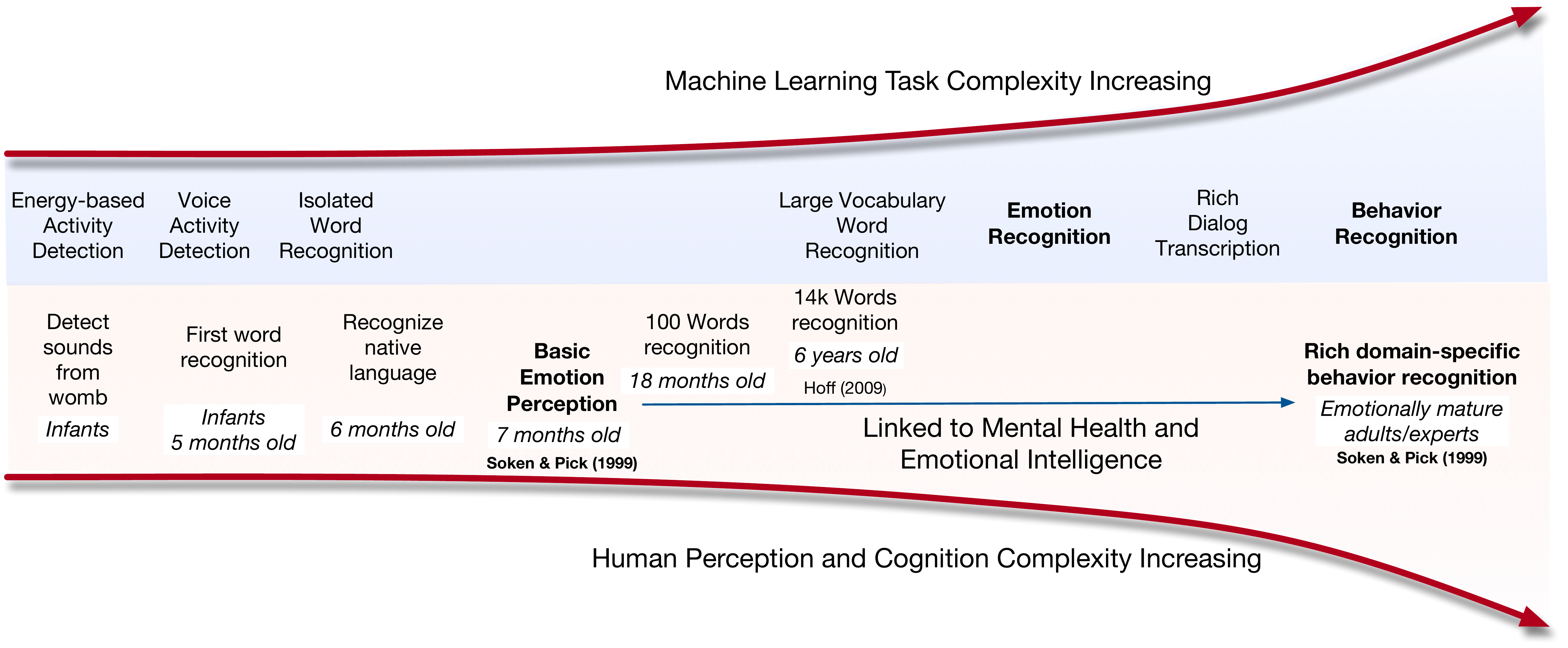}
    \caption{Illustration of task complexity or age of acquisition for machines and humans.}
    \label{fig:figure1}
\end{figure}

Those differences apply to both human cognition and machine learning aspects of speech capture, emotion recognition and behavior understanding as shown in Figure \ref{fig:figure1} \citep{soken1999infants, hoff2009language}. 
The increased complexity and contextualization of behavior can be seen both in humans as well as machines. For example, babies start to develop basic emotion perception at the age of seven months \citep{soken1999infants}. 
However, it takes emotionally mature and emotionally intelligent humans and often trained domain experts to perceive domain-specific behaviors.
In Figure \ref{fig:figure1}, we illustrate the complexity for machine processing along with the age-of-acquisition for humans. 
We see a parallel in the increase in demands of identifying behavior in both cases.

\paragraph{Motivations and goals of this work:}
\cm{knowledge hard, data driven} 
The relationship between emotion and behavior is usually implicit and highly nonlinear.
Investigating explicit and quantitative associations between behavior and emotions is thus challenging. 

\cm{Extension to my work}
In this work, based on the deep neural networks' (DNNs) underlying representative capability \citep{bengio2013representation, bengio2012deep}, we try to analyze and interpret the relationship between emotion and behavior information through data-driven methods.
We investigate the possibility of using transfer learning by employing emotion data as emotional related building blocks, or primitive features, that can describe behavior. Further, we design a deep learning framework that employs a hybrid network structure containing context dependent and reduced contextualization causality models to quantitatively analyze the relationship between basic emotions and complex behaviors.

\section*{Related Work} \zlabel{sec:related_work}
Researchers are combining machine learning techniques to build reliable and accurate emotion and behavior recognition systems.
Speech emotion recognition (SER) systems, of importance in human-computer interactions, enable agents and dialogue systems to act in a more human-like manner as conversational partners \citep{schuller2018speech}.
On the other hand, in the domain of behavior signal processing (BSP), efforts have been made in quantitatively understanding and modeling typical, atypical, and distressed human behavior with a specific focus on verbal and non-verbal communicative, affective, and social behaviors \citep{narayanan2013behavioral}.
We will briefly review the related work in the following aspects.

\paragraph{Emotion quantification from speech}\zlabel{sec:work_emo}
A dominant modality for emotion expression is speech \citep{cowie2003describing}.
Significant efforts \citep{el2011survey,beale2008affect,schuller2011recognising} have focused on automatic speech emotion recognition. 
Traditional emotion recognition systems usually rely on a two-stage approach, in which the feature extraction and classifier training are conducted separately. 
Recently, deep learning has demonstrated promise in emotion classification tasks \citep{han2014speech, le2013emotion}. 
Convolutional neural networks (CNNs) have been shown to be particularly effective in learning affective representations directly from speech spectral features \citep{mao2014learning, anand2015convoluted, huang2017characterizing, zheng2015experimental, aldeneh2017using}. 
\citet{mao2014learning} proposed to learn CNN filters on spectrally whitened spectrograms by an auto-encoder through unsupervised manners. 
\citet{aldeneh2017using} showed that CNNs can be directly applied to temporal low-level acoustic features to identify emotionally salient regions. 
\citet{anand2015convoluted} and \citet{huang2017characterizing} compared multiple kinds of convolutional kernel operations, and showed that the full-spectrum temporal convolution is more favorable for speech emotion recognition tasks. 
In addition, models with hidden Markov model (HMM) \citep{schuller2003hidden}, recurrent neural networks (RNNs) \citep{wollmer2010context,metallinou2012context,lee2015high} and the hybrid neural network combining CNNs and RNNs \citep{lim2016speech, huang2017deep} have also been employed to model emotion affect.

\paragraph{Behavior quantification from speech}\zlabel{sec:work_beh}
Behavioral signal processing (BSP) \citep{narayanan2013behavioral, georgiou2011behavioral} can play a central role in informing human assessment and decision making, especially in assisting domain specialists to observe, evaluate and identify domain-specific human behaviors exhibited over longer time scales. For example, in couples therapy \citep{black2013toward, nasir2017predicting}, depression \citep{gupta2014multimodal,nasir2016multimodal,stasak2016investigation,tanaka2017brain} and suicide risk assessment \citep{cummins2015review, venek2017adolescent, nasir2018_towards-an-unsu, nasir2017_complexity-in-s}, behavior analysis systems help psychologists observe and evaluate domain-specific behaviors during interactions. 
\citet{li2016_sparsely-connec} proposed sparsely connected and disjointly trained deep neural networks to deal with the low-resource data issue in behavior understanding. Unsupervised \citep{li2017unsupervised} and out-of-domain transfer learning \citep{tseng2018multi} have also been employed on behavior understanding tasks. 
Despite these important and encouraging steps towards behavior quantification, obstacles still remain. 
Due to the end-to-end nature of recent efforts, low-resource data becomes a dominant limitation \citep{li2016_sparsely-connec, collobert2011natural, soltau2017neural, heyman2001much}. 
This is exacerbated in BSP scenario by the difficulty of obtaining data due to privacy constraints \citep{lustgarten2015emerging, narayanan2013behavioral}. 
Challenges with subjectivity and low interannotator agreement \citep{busso2008expression,tseng2016_couples-behavio}, especially in micro and macro annotation complicate the learning task. 
Further, and importantly such end-to-end systems reduce interpretability generalizability and domain transfer \citep{sculley2015hidden}. 

\paragraph{Linking emotion and behavior quantification}\zlabel{sec:work_link}
As mentioned before, domain experts employ information within the context of the interaction at varying timescales to estimate the behaviors of interest \citep{narayanan2013behavioral, tseng2016_couples-behavio}. 
Specific short-term affect, \eg certain basic emotions, can be indicators of some complex long-term behaviors during manual annotation process \citep{heavey2002couples,jones1998couples, heyman2004rapid}.
These vary according to the behavior; for example, negativity is often associated with localized cues \citep{carney2007thin}, demand and withdrawal require more context \citep{heavey1995longitudinal}, and coercion requires a much longer context beyond a single interaction \citep{feinberg2007longitudinal}. \citet{sandeepJournal2019} analyzed behaviors, such as ``anger'' and ``satisfaction'', and found that negative behaviors could be quantified using short observation length whereas positive and problem solving behaviors required much longer observation. 

In addition, \citet{baumeister2007emotion, baumeister2010does} discussed two kinds of theories: the direct causality model and inner feedback model. Both models emphasize the existence of a relationship between basic emotion and complex behavior. 
Literature from psychology \citep{dunlop2008can, burum2007centrality} and social science \citep{spector2002emotion} also showed that emotion can have impacts and further shape certain complex human behaviors.
To connect basic emotion with more complex affective states, \citet{carrillo2016emotional} identified a relationship between emotional intensity and mood through lexical modality.
\citet{khorram2018priori} verified the significant correlation between predicted emotion and mood state for individuals with bipolar disorder on acoustic modality. All these indicate that the aggregation and link between basic emotions and complex behaviors is of interest and should be examined.

\section*{Proposed Work: Behavioral Primitives} \zlabel{sec:proposed_work}
Our work consists of three studies for estimation of behavior through emotion information as follows: 
\begin{enumerate}[noitemsep] 
\item \textbf{Context-dependent behavior from emotion labels:}
Basic emotion affect labels are directly used to predict long-term behavior labels through a recurrent neural network. This model is used to investigate whether the basic emotion states can be sufficient to infer behaviors. 
\item \textbf{Context-dependent behavior  from emotion-informed embeddings:}
Instead of directly using the basic emotion affect labels, we utilize emotion-informed embeddings towards the prediction of behaviors. 
\item \textbf{Reduced context-dependent behavior from emotion-informed embeddings:}
Similar to (2) above, we employ emotion-informed embeddings. In this case, however, we investigate the importance of context, by progressively reducing the context provided to the neural network in predicting behavior.
\end{enumerate}
For all three methods, we utilize a hybrid model of convolution and recurrent neural networks that we will describe in more detail below.

Through our work, both emotion labels and emotionally informed embeddings will be regarded as a type of behavior primitive, that we call \bpD. 

An important step in obtaining the above \bpS is the underlying emotion recognition system.
We thus first propose and train a robust \erD using convolutional neural network (CNN), which is described in detail in the following subsection. 

\subsection*{Emotion recognition} \zlabel{subsec:emotionRecognition}
In order to extract emotionally informed embeddings and labels, we propose a CNN based \erD.
The \erS model has a similar architecture as \citep{aldeneh2017using}, except that we use one-dimensional (1D) CNN kernels and train the network through a regression task. 
The CNN kernel filter should include entire spectrum information per scan, and shift along the temporal axis, which performs better than other kernel structures according to \citet{huang2017characterizing}.

Our model has three components: (1) stacked 1D convolutional layers; (2) an adaptive max pooling layer; (3) stacked dense layers.
The input acoustic features are first processed by multiple stacked 1D convolution layers. Filters with different weights are employed to extract different information from the same input sample.
Then, one adaptive max pooling layer is employed to further propagate 1D CNN outputs with the largest value.
This is further processed through dense layers to generate the emotional ratings at short-term segment level. 
The adaptive max pooling layer over time is one of the key components of this and all following models: First, it can cope with variable length signals and produce fixed size embeddings for subsequent dense layers; 
Second, it only returns the maximum feature within the sample to ensure only the more relevant emotionally salient information is propagated through training. 

\cm{explanation of our model}
We train this model as one regression model which predicts the annotation ratings of all emotions jointly. 
Analogous to the continuous emotion representation model \citep{schlosberg1954three}, this multi-emotion joint training framework can utilize strong bonds but blurred boundaries within emotions to learn the embeddings. 
Through this joint training process, the model can integrate the relationship across different emotions, and hopefully obtain an affective-rich embedding.

\cm{Structure of \erS}
In addition, to evaluate the performance of proposed \erS, we also build multiple binary, single-emotion, classification models (\ecD). 
The \ecS model is modified based on pre-trained \erS by replacing the last linear layer with new fully connected layers to classify each single emotion independently. 
During training, the back propagation only updates the newly added linear layers without changing the weights of pre-trained \erS model. 
In this case, the loss from different emotions is not entangled and the weights will be optimized towards each emotion separately. 
More details of experiments and results comparison are described in \namerefQ{Exp}.

\begin{figure}[ht]
    \centering
    \includegraphics[width=0.75\linewidth]{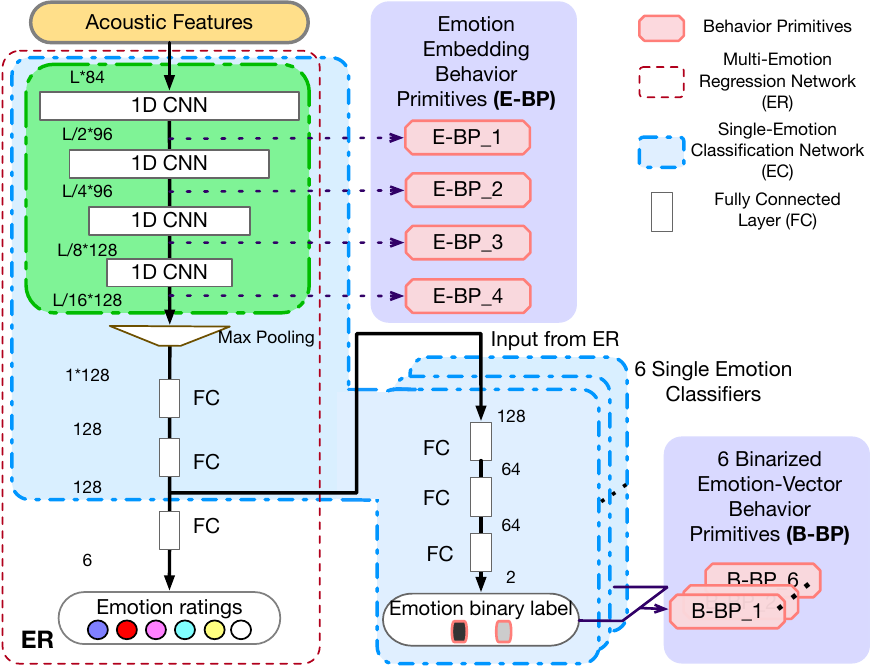}
    \caption{Models of ER, EC and two kinds of BPs. L is the input feature length.}
    \label{fig:figure2}
\end{figure}

\cm{Introduction of emotion ebds}As mentioned before, we employ \emph{two kinds of \bpM} in order to investigate the relationship between emotions and behaviors, and the selection of these two kinds of \bpS arises through the discrete, \ecS, and continuous, \erS, emotion representation models. 
The two kinds of \bpS are: (1) The discrete vector representation of predicted emotion labels, denoted as B-BP\_k, from the \ecD, where $k$ means $k^{\text{th}}$ basic emotion; and (2) The output embeddings of the CNN layers, denoted as E-BP\_$l$, from the \erD system, where $l$ represents the output from $l^{\text{th}}$ CNN layer.
All these are illustrated in Figure \ref{fig:figure2}. 

\subsection*{Behavior recognition through emotion-based behavior primitives}
We now describe three architectures for estimating behavior through \bpD.
The three methods employ full context of the emotion labels from the \ecD, the full context from the embeddings of the \erD system, and increasingly reduced context from the \erD system.

\subsubsection*{Context-dependent behavior recognition from emotion labels} \zlabel{subsubsec:behRecognition_seq}
In this approach, the binarized predicted labels from the \ecS system are employed to predict long-term behaviors via sequential models in order to investigate relationships between emotions and behaviors. 
Such a design can inform the degree to which short-term emotion can influence behaviors.
It can also provide some interpretability of the employed information for decision making, over end-to-end systems that generate predictions directly from the audio features.

\begin{figure}[ht]
    \centering
    \includegraphics[width=0.8\linewidth]{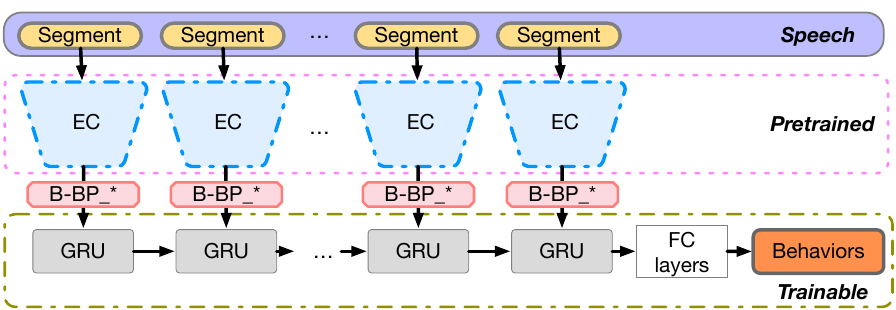}
    \caption{\ecbpS based context-dependent behavior recognition model}
    \label{fig:figure3}
\end{figure}

\cm{model structure}
We utilize the \ecD described in the previous section to obtain the predicted \ecbpD  on shorter speech segment windows as \bpM. 
These are extracted from the longer signals that describe the behavioral corpus and are utilized, preserving sequence, hence context, within a recurrent neural network for predicting the behavior labels.
Figure \ref{fig:figure3} illustrates the network architecture and B-BP\_* means the concatenation of all B-BP\_k, where k ranges from 1 to 6.

In short, the \ecbpS vectors are fed into a stack of gated recurrent units (GRUs), followed by a densely connected layer which maps the last hidden state of the top recurrent layer to behavior label outputs. 
GRUs were introduced in \citet{chung2014empirical} as one attempt to alleviate the issue of vanishing gradient in standard vanilla recurrent neural networks and to reduce the number of parameters over long short-term memory (LSTM) neurons.
GRUs have a linear shortcut through timesteps which avoids the decay and thus promotes gradient flow. 
In this model, only the sequential GRU components and subsequent dense layers are trainable, while the \ecS networks remain fixed.

\subsubsection*{Context-dependent behavior recognition from emotion-embeddings} \zlabel{subsubsec:e2b-embed}

It is widely understood that information closer to the output layer is more tied to the output labels while closer to the input layer information is less constrained and contains more information about the input signals.
In our \erS network, the closer we are to the output, the more raw information included in the signal is removed and the more we are constrained to the basic emotions.
Given that we are not directly interested in the emotion labels, but in employing such relevant information for behavior, it makes sense to employ layers below the last output layer to capture more behavior-relevant information closer to its raw form.
Thus, instead of using the binary values representing the absence or existence of the basic emotions,  we can instead employ \erbpD \xspace as the input representation.

\begin{figure}[ht]
    \centering
    \includegraphics[width=0.95\linewidth]{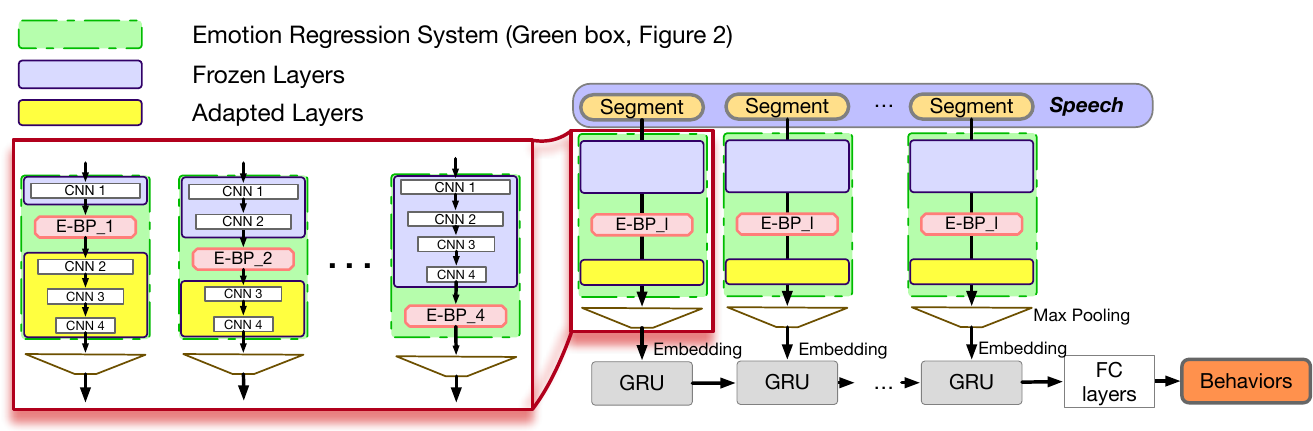}
    \caption{\erbpS based context-dependent behavior recognition model. E-BP\_$l$ is the output from $l^{\text{th}}$ pretrained CNN layer. In practice multiple E-BP\_$l$ can be employed at the same time through concatenation. In this work we only employ the output of a single layer at a time.}
    \label{fig:figure4}
\end{figure}

The structure of the system is illustrated in Figure \ref{fig:figure4}. 
After pretraining the \erS, we keep some layers of that system fixed, and employ their embeddings as the \erbpD. 
We will discuss the number of fixed layers in the experiments section.
This \erbpS serves as the input of the subsequent, trainable, convolutional and recurrent networks. 

The overall system is trained to predict the longer-term behavior states. 
By varying the number of layers that remain unchanged in the \erS system and using different embeddings from different layers for the behavior recognition task we can identify the best embeddings under the same overall number of parameters and network architecture.

The motivation of the above is that the fixed \erS encoding module is focusing on learning emotional affect information, which can be related but not directly linked with behaviors. 
By not using the final layer, we are employing a more raw form of the emotion-related information, without extreme information reduction, that allows for more flexibility in learning by the subsequent behavior recognition network.
This allows for transfer learning \citep{torrey2010transfer} from one domain (emotions) to another related domain (behaviors). 
Thus, this model investigates the possibility of using transfer learning by employing emotional information as ``building blocks'' to describe behavior.


\subsubsection*{Reduced context-dependent behavior recognition from emotion-informed embeddings} \zlabel{subsec:e2b-reduced_context}

\cm{intro, seq vs. non-seq}
In the above work, we assume that the sequence of the behavior indicators (embeddings or emotions) is important.
To verify the need for such an assumption, in this section, we propose varying the degree of employed context. 
Through quantification, we analyze the time-scales at which the amount of sequential context affects the estimation of the underlying behavioral states.

\begin{figure}[ht]
    \centering
    \includegraphics[width=0.5\linewidth]{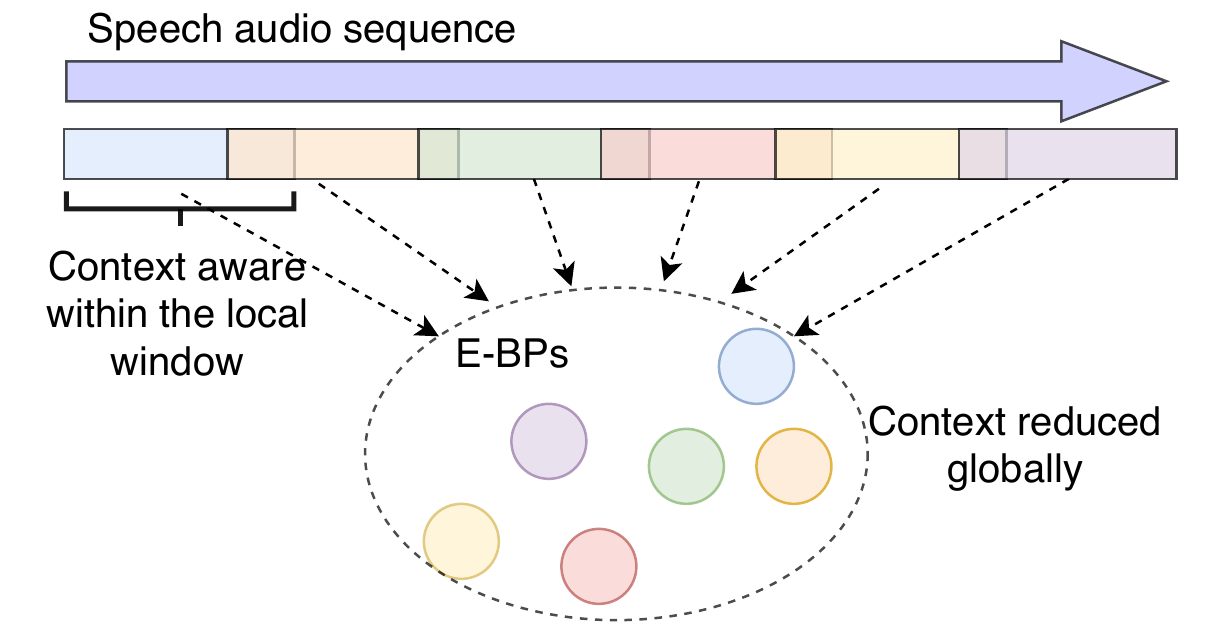}
    \caption{Illustration of local context awareness and global context reduction. In previous sections, the E-BPs (and E-BPs) are passed to a GRU that preserves their sequences. Here they are processed through pooling and context is removed.}
    \label{fig:figure5}
\end{figure}

In this proposed model, we design a network that can only preserve local context.
The overall order of the embeddings extracted from the different local segments is purposefully ignored so we can better identify the impact of de-contextualizing information as shown in Figure \ref{fig:figure5}.

\begin{figure}[ht]
    \centering
    \includegraphics[width=0.7\linewidth]{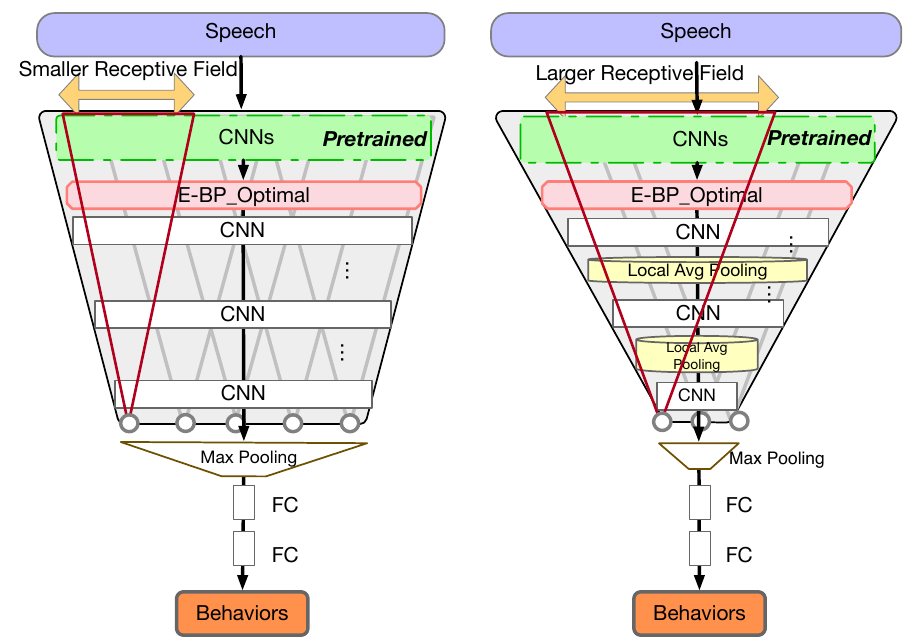}
    \caption{\erbpS reduced context-dependent behavior recognition model}
    \label{fig:figure6}
\end{figure}

In practice, this reduced-context model is built upon the existing CNN layers as in the \erbpS case. 
We will create this reduced context system by employing only the \erbpS embeddings. The \erbpS embeddings are extracted from the same emotion system as before.
In this case, however, instead of being fed to a recursive layer with full-session view, 
we eliminate the recursive layer and incorporate a variable number of CNN layers and local average pooling functions in between to adjust context view. 
Since the final max-pooling layer ignores the order of the input, the largest context is determined by the receptive field view of the last layer before this max-pooling. 
We can thus investigate the impact of context by varying the length of the CNN receptive field.

Figure \ref{fig:figure6} illustrates the model architecture. We extract the optimal \erbpS based on the results of previous model, and then employ more CNN layers with different receptive field sizes to extract high-dimensional representation embeddings, and finally input them to the adaptive max-pooling along the time axis to eliminate the sequential information.
Within each CNN receptive field, shown as red triangles in the figure, the model still has access to the full receptive field context.
The max pooling layer removes context across the different receptive windows.

Furthermore, the receptive field can be large enough to enable the model to capture behavioral information encoded over longer timescales.
In contrast a very small receptive area, \eg at timescale of phoneme or word, sensing behaviors should be extremely difficult \citep{baumeister2010does} and can even be challenging to detect emotions \citep{mower2011hierarchical}.
The size of the receptive field is decided by the number of CNN layers, corresponding stride size, and the number of local average pooling layers in between. 
In our model, we adjust the size of the receptive field by setting different number of local average pooling layers under which the overall number of network parameters is unchanged.

\section*{Datasets} \zlabel{database}
\subsection*{Emotion dataset: CMU-MOSEI Dataset}
The CMU Multimodal Opinion Sentiment and Emotion Intensity (CMU-MOSEI) \citep{zadeh2018multimodal} contains video files carefully chosen from YouTube.
Each sample is a monologue with verified quality video and transcript. 
This database includes 1000 distinct speakers with 1000 kinds of topics, and are gender balanced with an average length of 7.28 seconds. 
Speech segments are already filtered during the data collection process, thus all speech segments are monologues of verified audio quality.

For each speech segment, six emotions (Happiness, Sadness, Anger, Fear, Disgust, Surprise) are annotated on a [0,3] Likert scale for the presence of each emotion. (0: no evidence; 1: weak evidence; 2: evidence; and 3: high evidence of emotion). 
This, after averaging ratings from 3 annotators, results in a 6-dimensional emotional rating vector per speech segment. 
CMU-MOSEI ratings can also be binarized for each emotion: if a rating is greater than 0 it is considered that there is some presence of emotion, hence it is given a true presence label, while a zero results in a false presence of the emotion.

The original dataset has 23,453 speech segments and each speech segment may contain more than one emotion presence label. 
Through our experiments, we use the segments with available emotion annotations and standard speaker independent split from dataset SDK \citep{a2zadeh_2019}: Overall we have true presence in 12465 segments for happiness, 5998 for sadness, 4997 for anger, 2320 for surprise, 4097 for disgust and 1913 for fear. 
Due to the imbalance, accurate estimation of some emotions will be challenging.
The training set consists of 16331 speech segments, while the validation set and test set consist of 1871 and 4662 sentences respectively. 

\subsection*{Behavior dataset: Couples Therapy Corpus}\zlabel{database:couple}
The Couples Therapy dataset is employed to evaluate complex human behaviors. The corpus was collected by researchers from the University of California, Los Angeles and the University of Washington for the Couple Therapy Research Project \citep{christensen2004traditional}.
It includes a longitudinal study of 2 years of 134 real distressed couples.
Each couple has been recorded at multiple instances over the 2 years.
At the beginning of each session, a relationship-related topic (e.g. ``why can't you leave my stuff alone?'') was selected and the couple interacted about this topic for 10 minutes. 
Each participant's behaviors were rated by multiple well-trained human annotators based on the Couples Interaction \citep{heavey2002couples} and Social Support Interaction \citep{jones1998couples} Rating Systems. 
31 behavioral codes were rated on a Likert scale of 1 to 9, where 1 refers absence of the given behavior and 9 indicates a strong presence.
Most of the sessions have 3 to 4 annotators, and annotator ratings were averaged to obtain the final 33-dimensional behavioral rating vector.
The employed part of the dataset includes 569 coded sessions, totaling 95.8 hours of data across 117 unique couples.

\section*{Audio processing and feature extraction}\zlabel{audio feature}
\subsection*{Behavioral dataset pre-processing} \label{audio_processing}
For preprocessing the couples therapy corpus we employ the procedure described in \citep{black2013toward}.
The main steps are Speech Activity Detection (SAD) and diarization.
Since we only focus on acoustic features extracted for speech regions, we extract the speech parts using the SAD system described in \citep{ghosh2011robust}, and only keep sessions with an average SNR greater than 5 dB (72.9\% of original dataset). 
Since labels of behavior are provided per-speaker, accurate diarization is important in this task. 
Thus, for diarization we employ the manually-transcribed sessions and a forced aligner in order to achieve high quality interlocutor-to-audio alignment.
This is done using the recursive ASR-based procedure of alignment of the transcripts with audio by \emph{SailAlign} \citep{katsamanis2011sailalign}. 

Speech segments from each session for the same speaker are then used to analyze behaviors. 
During testing phase, a leave-test-couples-out process is employed to ensure separation of speaker, dyad, and interaction topics. More details of the preprocessing steps can be found in \citep{black2013toward}. 

After the processing procedure above, the resulting corpus has a total of 48.5 hours of audio data across 103 unique couples and a total of 366 sessions.

\subsection*{Feature extraction} \zlabel{feat_extraction}
In this work, we focus only on the acoustic features of speech. 
We utilize Log-Mel filterbank energies (Log-MFBs) and MFCCs as spectrogram features.
Further, we employ pitch and energy. 
These have been shown in past work to be the most important features in emotion and behavior related tasks. 
These features are extracted using Kaldi \citep{povey2011kaldi} toolkit with a 25 ms analysis window and a window shift of 10 ms. 
The number of Mel-frequency filterbanks and MFCCS are both set to 40. 
For pitch, we use the extraction method in \citep{ghahremani2014pitch}, in which 3 features, normalized cross correlation function (NCCF), pitch ($f_0$), the delta of pitch, are included for each frame. 

After feature extraction, we obtain an 84-dimensional feature per frame (40 log-MFB's, 40 MFCC's, energy, $f_0$, delta of $f_0$, and NCFF). 

\section*{Experiments and Results Discussion}\zlabel{Exp}
\subsection{General settings}

For emotion-related tasks, we utilize the CMU-MOSEI dataset with the given standard train, validation, test data split from \citep{a2zadeh_2019}. 

For the behavior related tasks,  we employ the couple therapy corpus and use leave-4-couples-out cross-validation. 
Note that this results in 26 distinct neural-network training-evaluation cycles for each experiment.
During each fold training, we randomly split 10 couples out as a validation dataset to guide the selection of the best trained model and prevent overfitting. 
All these settings ensure that the behavior model is speaker independent and will not be biased by speaker characteristics or recording and channel conditions.

\begin{table}[ht]
\scalebox{0.95}{
\begin{tabular}{c|c}
\hline
Behavior & Description \\ \hline
Acceptance         &  Indicates understanding, acceptance, respect for partner's views, feelings and behaviors           \\ 
Blame         &  Blames, accuses, criticizes partner and uses critical sarcasm and character assassinations           \\ 
Positivity         & Overtly expresses warmth, support, acceptance, affection, positive negotiation\\ 
Negativity         &  Overtly expresses rejection, defensiveness, blaming, and anger           \\ 
Sadness         &   Cries, sighs, speaks in a soft or low tone, expresses unhappiness and disappointment          \\ \hline
\end{tabular}
}
\caption{Description of behaviors}
\label{tab:table1}
\end{table}

In our experiments, we employ five behavioral codes: \emph{Acceptance, Blame, Positivity, Negativity} and \emph{Sadness}, each describing a single interlocutor in each interaction of the couples therapy corpus. 
Table \ref{tab:table1} lists a brief description\footnote{Full definitions are too long to insert in this manuscript and reader is encouraged to look into \citep{heavey2002couples, jones1998couples}} of these behaviors from the annotation manuals \citep{heavey2002couples,jones1998couples} .

Following the same setting of \citep{black2010automatic} to reduce effects of interannotator disagreement, we model the task as a binary classification task of low- and high- presence of each behavior. This also enables balancing for each behavior resulting in equal-sized classes. This is especially useful as some of the classes, \eg Sadness, have an extremely skewed distribution towards low ratings. 
More information on the distribution of the data and impact on classification can be found in \citep{georgiou2011aggravating}.
Thus, for each behavior code and each gender, we filter out 70 sessions on one extreme of the code (e.g., high blame) and 70 sessions at the other extreme (e.g., low blame). 

\cm{how to calculate loss}
Since due to the data cleaning process, some sessions may be missing some of the behavior codes, we use a mask and train only for the available behaviors.
Moreover, the models are trained to predict the binary behavior labels for all behaviors together. The loss is calculated by averaging 5 behavioral classification loss with masked labels. 
Thus, this loss is not optimizing for any specific behavior but it is focusing on the general, latent, link between emotions and behaviors. 

\subsection*{\erS and \ecS for Emotion Recognition}\zlabel{exp:speech_emo_reco}

\cm{model details}Both the \erD  and the \ecD are trained using the CMU-MOSEI dataset. 

The \erD system consists of 4 layers of 1D CNN layers, adaptive max-pooling layer and followed by 3 fully connected layers with ReLU activation function. 
During the training, we randomly choose a segment from each utterance and represent the label of the segment using the utterance label.
In our work, we employ a segment length of 1 second.

The model is trained jointly with all six emotions by optimizing the mean square error (MSE) regression loss for all emotions ratings together using Adam optimizer \citep{kingma2014adam}.

In a stand-alone emotion regression task, a separate network that can optimize per-emotion may be needed (through higher-level disconnected network branches), however in our work, as hypothesized above, this is not necessary. 
Our goal is to extract as much information as possible from the signal relating to any and all available emotions.
We will, however, investigate optimizing per emotion in the \ecS case.

\cm{emotion binary recognition exp}
Further to the \erS system, we can optimize per emotion through the \ecD. 
This is trained for each emotion separately by replacing the pre-trained \erS's last linear layer with three emotion-specific fully connected layers. 
We use the same binary labeling setting as described in \citep{zadeh2018multimodal}: Within each emotion, for samples with original rating value larger than zero, we assign the label 1 by considering the presence of that emotion; for samples with rating 0, we assign label 0. 
During training, we randomly choose 1-second segments as before. 
During evaluation, we segment each utterance into one-second segments and the final utterance emotion label is obtained via majority voting.
In addition, the CMU-MOSEI dataset has a significant data imbalance issue: the true label in each emotion is highly under-represented. To alleviate this, during training, we balance the two classes by subsampling the 0 label esence class in every batch.

In our experiments, in order to correctly classify most of the relevant samples, the model is optimized and selected based on average weighted accuracy (WA) as used in \citep{zadeh2018multimodal}. WA is defined in \citep{tong2017combating}: Weighted Accuracy = $(TP \times N/P + TN)/2N$, where $TP$ (resp. $TN$) is true positive (resp. true negative) predictions, and $P$ (resp. $N$) is the total number of positive (resp. negative) examples. 
As shown in Table \ref{tab:table2}, we present WA of each \ecS system and compare them with the state-of-art results from \citep{zadeh2018multimodal}. 

\begin{table}[ht]
\centering
\scalebox{0.9}{
\begin{tabular}{ccccccc}
\hline
Emotions          & Anger     & Disgust   & Fear      & Happy     & Sad  & Surprise  \\ \hline

Methods in CMU-MOSEI\\ \citep{zadeh2018multimodal} & 56.4   & 60.9   & \textbf{62.7}  & 61.5     & 62.0    & 54.3  \\ \hline
Proposed \ecS & \textbf{61.2}  & \textbf{64.9}  &  57.0                 & \textbf{63.1}  & \textbf{62.5}     & \textbf{56.2}         \\ \hline
\end{tabular}
}
\caption{Weighted classification accuracy (WA) in percentage for emotion recognition on the CMU-MOSEI dataset. Bold numbers represent the best performing system.}
\label{tab:table2}
\end{table}

Compared with \citep{zadeh2018multimodal}, our proposed 1D CNN based emotion recognition system achieves comparable results and thus the predicted binary emotion labels can be considered satisfactory for further experiments. 
More importantly, our results indicate that the pre-trained \erS embedding captures sufficient emotion related information and can thus be employed as a  behavior primitive.

\subsection*{Context-dependent behavior recognition} \label{ext:exp_seq_beh}
The main purpose of the experiments in this subsection is to verify the relationship between emotion-related primitives and behavioral constructs. We employ both \ecbpS and \erbpS as described below. Before that, we first use examples to illustrate the importance of context information in behavior understanding. 

\subsubsection*{Importance of context information in behavior understanding} \label{exp:inportance_context}
\begin{figure}[ht]
    \centering
    \includegraphics[width=0.75\linewidth]{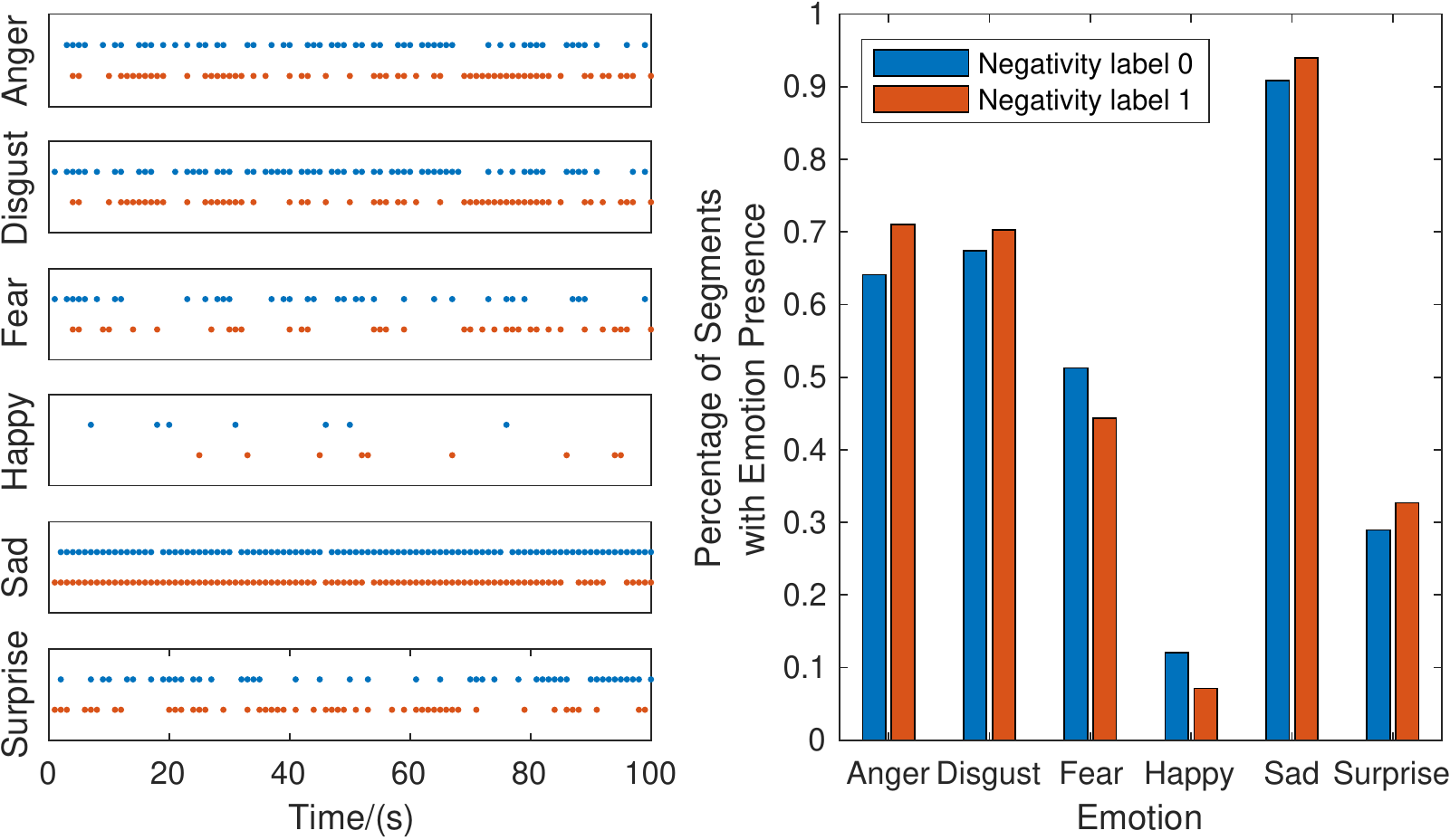}
    \caption{Sessions with similar percentage of emotions presence but different behavior label}
    \label{fig:figure7}
\end{figure}

Prior to presenting the behavior classification results, we use two sessions from couple therapy corpus to illustrate the importance of context information in behavior understanding. 
Once the \ecD systems are trained, a sequence of emotion label vectors can be generated by applying the \ecS systems on each speech session. 
We choose two sessions and plot those sequences of emotion presence vectors of the first 100 seconds as an example in Figure \ref{fig:figure7}, in which each dot represents the emotion presence (\ie predicted label equals to 1) at the corresponding time. 
For each emotion, the percentage of emotion presence segments is calculated by dividing the number of emotion presence segments by the total number of segments.

These two sessions are selected as an example since they have similar audio stream length and percentage of emotion presence segments but different behavior labels: the red represents one session with ``strong presence of negativity'' while blue represents another session with ``absence of negativity''. 
This example reveals the fact that, as we expected, the behaviors are determined not only by the percentage of affective constructs but also the contextual information. 
As shown in the left side of Figure \ref{fig:figure7}, the emotion presence vectors exhibit different sequential patterns within two sessions, even though no significant distribution difference can be observed.

\subsubsection*{\ecbpS based context-dependent behavior recognition} \label{exp:exp_labels2beh}
\ecbpD are generated by applying the \ecD systems on the couple therapy data: 
For each session, a sequence of emotion label vectors is generated as $\emph{E} = [\mathbf{e_1}, \mathbf{e_2}, ..., \mathbf{e_T}]$, where each element $\mathbf{e_i}$ is the 6 dimensional \ecbpS binary label vector at time $i$. 
That means that ${e_{ij}}$ represents the presence, through a binary label 0 or 1, of emotion $j$ at time $i$. 
Such \ecbpS are the input of the context-dependent behavior recognition model that has two layers of GRUs followed by two linear layers as illustrated in Fig. \ref{fig:figure3}. 

\begin{table}[ht]
\centering
\begin{tabular}{cccccc}
\hline
 \textbf{Average} & Acceptance & Blame & Positivity & Negativity & Sadness \\ \hline
 60.43            & 61.07      & 63.21 & 59.64      & 59.29      & 58.93   \\ \hline
\end{tabular}
\caption{Behavior binary classification accuracy in percentage for context-dependent behavior recognition model from emotion labels}
\label{tab:table3}
\end{table}

As shown in Table \ref{tab:table3}, the average binary classification accuracy of these five behaviors is 60.43\%. 
Considering that the classification accuracy can reach up to 50\% by chance with balanced data, our results show that behavioral states can be weakly inferred from the emotion label vector sequences.
Further, we perform the McNemar test, and the results above and throughout the paper are statistically significant with $p<0.01$.
Despite the low accuracy of the behavior positivity, these results suggest a relationship between emotions and behaviors that we investigate further below. 

\subsubsection*{\erbpS based context-dependent behavior recognition}\zlabel{exp:exp_ebds2beh}
The simple binary emotion vectors (as \ecbpS) indeed link emotions and behaviors. 
However, they also demonstrate that the binarized form of \ecbpS limits the provided information bandwidth to higher layers in the network, and as such limits the ability to predict the much more complex behaviors. 
These are reflected in the low accuracies in Table \ref{tab:table3} . 

This further motivates the use of the \erbpD.
As described in Figure \ref{fig:figure4}, we construct input of the \erbpS context-dependent behavior recognition system using the pretrained \erD. 
These \erbpS embeddings capture more information than just the binary emotion labels. 
They potentially capture a higher abstraction of emotional content, richer paralinguistic information, conveyed through a non-binarized version that doesn't limit the information bandwidth, and may further capture other information such as speaker characteristics or even channel information. 

We employ embeddings from different layers of the \erS network. The layers before the employed embedding are in each case frozen and only the subsequent layers are trained as denoted in Figure \ref{fig:figure4}. 
The trainable part of the network includes several CNN layers with max pooling and subsequent GRU networks. 
The GRU part of the network is identical to the ones used by the context-dependent behavior recognition from \erbpS.

The use of different depth embeddings can help identify where information loss becomes too specific to the \erS loss objective versus where there is too much unrelated information to the behavior task. 

\begin{table}[ht]
\centering
\scalebox{0.9}{
\begin{tabular}{ccccccc}
\hline
 & \textbf{Average} & Acceptance & Blame & Positivity & Negativity & Sadness \\ \hline
\begin{tabular}[c]{@{}c@{}}None-\erbpS model\\ (Baseline)\end{tabular} 
                & 58.86            & 62.86               & 62.50              & 57.86           & 60.00            & 51.07   \\ \hline
E-BP\_1 model   & 59.79            & 64.29               & 62.86              & 60.00           & 61.07      & 50.71   \\ \hline
E-BP\_2 model  & 60.79            & 61.79                & 63.93              & 62.86           & 63.57            & 51.79   \\ \hline
E-BP\_3 model  & 65.00            & 66.07                & 69.29              & \textbf{65.36}   & 69.29             & 55.00   \\ \hline
E-BP\_4 model  & \textbf{69.00}            & \textbf{72.50}      & \textbf{71.79}     & \textbf{65.36}    &  \textbf{76.07}   & \textbf{59.29}   \\ \hline
\end{tabular}
}
\caption{Behavior binary classification accuracy in percentage for context-dependent behavior recognition model from emotion-embeddings. Bold numbers represent the best performing system.}
\label{tab:table4}
\end{table}

In Table \ref{tab:table4}, the none-\erbpS model, as the baseline, means all parameters are trained from random initialization instead of using the pretrained \erbpS input.
While E-BP\_$l$ model means the first $l$ layers of the pretrained \erS network are fixed and their output is used as the embedding \erbpS for the subsequent system. 
As seen in the second column of the table, all of \erbpS based models perform significantly better than the \ecbpS based model, which achieves an improvement of 8.57\% on average and up to 16.78\% for Negativity. 

These results, further support the use of basic emotions as constructs of behavior. 
In general, for all behaviors, the higher-level \erbpS s, which are closer to the \erS loss function, can capture affective information and obtain better performance in behavior quantification compared with lower-level embeddings. 
From the description in Table \ref{tab:table1}, some behaviors are closely related to emotions. 
For example, negativity is defined in part as "Overtly expresses rejection, defensiveness, blaming, and anger", and sadness\footnote{Which isn't necessarily perfectly aligning with the basic emotion "sad" but follows the SSIRS manual} is defined in part as "expresses unhappiness and disappointment". This shows that these behaviors are very related to emotions such as anger and sad, thus it's expected that an embedding closer to the \erS loss function will behave better. Note that these are not at all the same though: a negative behavior may mean that somewhere within the 10 min interaction or through unlocalized gestalt information the expert annotators perceived negativity; in contrast a negative emotion has short-term information (on average 7s segment) that is negative.

An interesting experiment is what happens if we use a lower-ratio of emotion (out-of-domain) vs. behavior (couples-in-domain) data.
To perform this experiment we use only half of the CMU-MOSEI data\footnote{11875 samples from commit: \\ https://github.com/A2Zadeh/CMU-MultimodalSDK/commit/f0159144f528380898df8093381c8d83fd7cc475} to train another \erS system, and use this less robust \erS system and corresponding \erbpS representations to reproduce the behavior quantification as in Table \ref{tab:table4}. 
What we observe is that the reduced learning taking place on emotional data requires the in-domain system to have prefer embeddings closer to the feature. Specifically Negativity performs equally well with layers 3 or 4 at 71.43\%. Positivity performs best with layer 3 at 64.64\%, Blame and Acceptance perform best with layer 2 at 71.07\% and 72.86\% respectively while Sadness performs best through layer 1 at 56.07\%. 

In the reduced data case we observe that best performing layer is not consistently layer 4. Employing the full dataset as in Table \ref{tab:table4} provides better performance than using less data and in that case layer 4 (E-BP\_4) is always the best performing layer, thus showing that more emotion data provides better ability of transfer learning.

\subsection*{Reduced context-dependent behavior recognition}\zlabel{exp:seq_indep}

In the previous two sections we demonstrate that there is a benefit to transfer emotion-related knowledge to behavior tasks. 
We show that the wider bandwidth information transfer through an embedding \erbpS is beneficial to a binarized \ecbpS representation. 
We also show that depending on the degree of relationship of the desired behavior to the signal or to the basic emotion, different layers that are closer to the input signal or closer to the output loss, may be more or less appropriate. 
However, in all the above cases we assume that the sequence and contextualization of the extracted emotion information was needed. That is captured and encoded through the recursive GRU layers.

We conduct an alternative investigation into how much contextual information is needed. 
As discussed in section \namerefQ{subsec:e2b-reduced_context} and shown on Figure \ref{fig:figure6} we can reduce context through changing the receptive field of our network prior to removing sequential information via max pooling.

In this section we select the best \erbpS based on average results in Table \ref{tab:table4}, \ie E-BP-4, as the input of the reduced context-dependent behavior recognition model.
Based on E-BP-4 embeddings, the reduced context-dependent model employs 4 more CNN layers with optional local average pooling layers in between, and is followed by an adaptive max pooling layer and three fully connected layers to predict the session level label directly without sequential modules. 

Since the number of parameters of this model is largely increased, dropout \citep{srivastava2014dropout} layers are also utilized to prevent overfitting.
Local average pooling layers with kernel size 2 and stride 2 are optionally added between newly added CNN layers to adjust the final size of the receptive field: The more average pooling layers we use, the larger temporal receptive field can be obtained for the same number of network parameters.
We endure that the overall number of trainable parameters is the same for the different receptive field settings, which provides a fair comparison of the resulting systems.
The output of these CNN/local pooling layers is passed to an adaptive max pooling before the fully connected layers as in Figure \ref{fig:figure6}.

\begin{table}[htb]
\centering
\scalebox{0.9}{
\begin{tabular}{ccccccc}
\hline
                         & \textbf{Average} & Acceptance & Blame       & Positivity & Negativity & Sadness \\ \hline
Receptive\_field\_ 4s & 63.43            & 65.00      & 70.00       & 58.92      & 67.50      & 55.71   \\ \hline
Receptive\_field\_ 8s  & 62.71            & 65.00      & 69.64       & 56.79     & 66.07      & 56.07   \\ \hline
Receptive\_field\_ 16s   & 63.36            & 63.57      & 69.64       & 60.71     & 66.42      & \textbf{56.43}   \\ \hline
Receptive\_field\_ 32s   & \textbf{66.36} &\textbf{68.21}& \textbf{73.21}& \textbf{63.21}  & 71.43 & 55.71   \\ \hline
Receptive\_field\_ 64s   &65.57   & 66.43     & 72.86        & 62.50        & \textbf{71.79}       & 54.29   \\ \hline
\end{tabular}
}
\caption{Behavior binary classification accuracy in percentage for reduced context-dependent behavior recognition from emotion-informed embeddings. Bold numbers represent the best performing system.}
\label{tab:table5}
\end{table}

In Table \ref{tab:table5}, each model has a different temporal receptive window ranging from 4 seconds to 1 minute. 
For most behaviors, we observe a better classification as the receptive field size increases, especially in the range from 4 seconds to 32 seconds, demonstrating a need for longer observations for behaviors.

Furthermore, the results suggest different behaviors require different observation window length to be quantified, which is also observed by \citet{sandeepJournal2019} using lexical analysis. By comparing results with different receptive window sizes, we can indirectly obtain the appropriate behavior analysis window size for each behavior code.
As shown in Table \ref{tab:table5}, sadness has a smaller optimal receptive field size than behaviors such as acceptance, positivity and blame. 
This is in good agreement with the behavior descriptions. 
For example, behaviors of acceptance, positivity and blame often require relatively longer observations since they relate to understanding and respect for partner's views, positive negotiation, and accusation respectively, which often require multiple turns in a dialog and context to be captured.
On the other hand, sadness which can be expressed via emitting a long, deep, audible breath, and is also related to short-term expression of unhappy affect, can be captured with shorter windows. 

Moreover, we find the classification of negativity reaches high accuracy when using a large receptive field. 
This might be contributed by the fact that the negative behavior in the couple therapy domain is complex, which is not only revealed by short term negative affect but also related to context based negotiation and hostility, and is captured through gestalt perception of the interaction.

In addition, the conclusion that most of the behaviors do not benefit much from longer than 30 seconds\footnote{Note that this does not make any claims on interlocutor dynamics, talk time, turn-taking \etc, but just single person acoustics} windows matched existing literature on thin slices \citep{ambady1992thin}, which refer to excerpts of an interaction that can be used to arrive at a similar judgment of behavior to as if the entire interaction had been used.

\subsection*{Analysis on behavior prediction uncertainty reduction} \zlabel{metrics}

Besides the verification of the improvement from \ecbpS based model to \erbpS based models, in this section, we further analyze the importance of context information for each behavior by comparing results between \erbpS based context-dependent and reduced context-dependent models. 
This analysis calls into question that which behavior is more context involved and to what degree.

Classification accuracy is used as the evaluation criterion in previous experiments. More generally, this number can be regarded as a probability of correct classification when a new session comes to measure. 
Inspired by entropy from information theory, we define one metric named Prediction Uncertainty Reduction (PUR) and use it to indicate the relative behavior prediction and interpretation improvement among different models for each behavior.

Suppose $p_{m}(x) \in [0,1]$ is the probability of correct classification for behavior $x$ with model $m$.
We define the uncertainty of behavior prediction as:
$$I_{m}(x) = -p_{m}(x)log_2(p_{m}(x))-(1-p_{m}(x))log_2(1-p_{m}(x))$$
if $p_{m}(x)$ is equal to 1, $I_{m}(x) = 0$ there is no improvement possibility; if $p_{m}(x)$ is equal to 0.5, same as random prediction accuracy, the uncertainty is the largest.
We further define the Prediction Uncertainty Reduction (PUR) value of behavior $x$ from model $m$ to model $n$ as: $$R_{m \rightarrow n}(x)= I_{m}(x) - I_{n}(x)$$
We use this value to indicate improvements between different models. 

We use PUR to sense the relative improvement from \erbpS based context-dependent and \erbpS based reduced context-dependent models respectively, to the baseline \ecbpS based context-dependent model.
The larger value of PUR suggests the clear improvement of behavior prediction.
For each behavior, for each \erbpS based model, we choose the best performance model (the bold number from Table \ref{tab:table4} and Table \ref{tab:table5}) to calculate PUR value from baseline \ecbpS context-dependent model.

\begin{figure}[ht]
    \centering
    \includegraphics[width=0.6\linewidth]{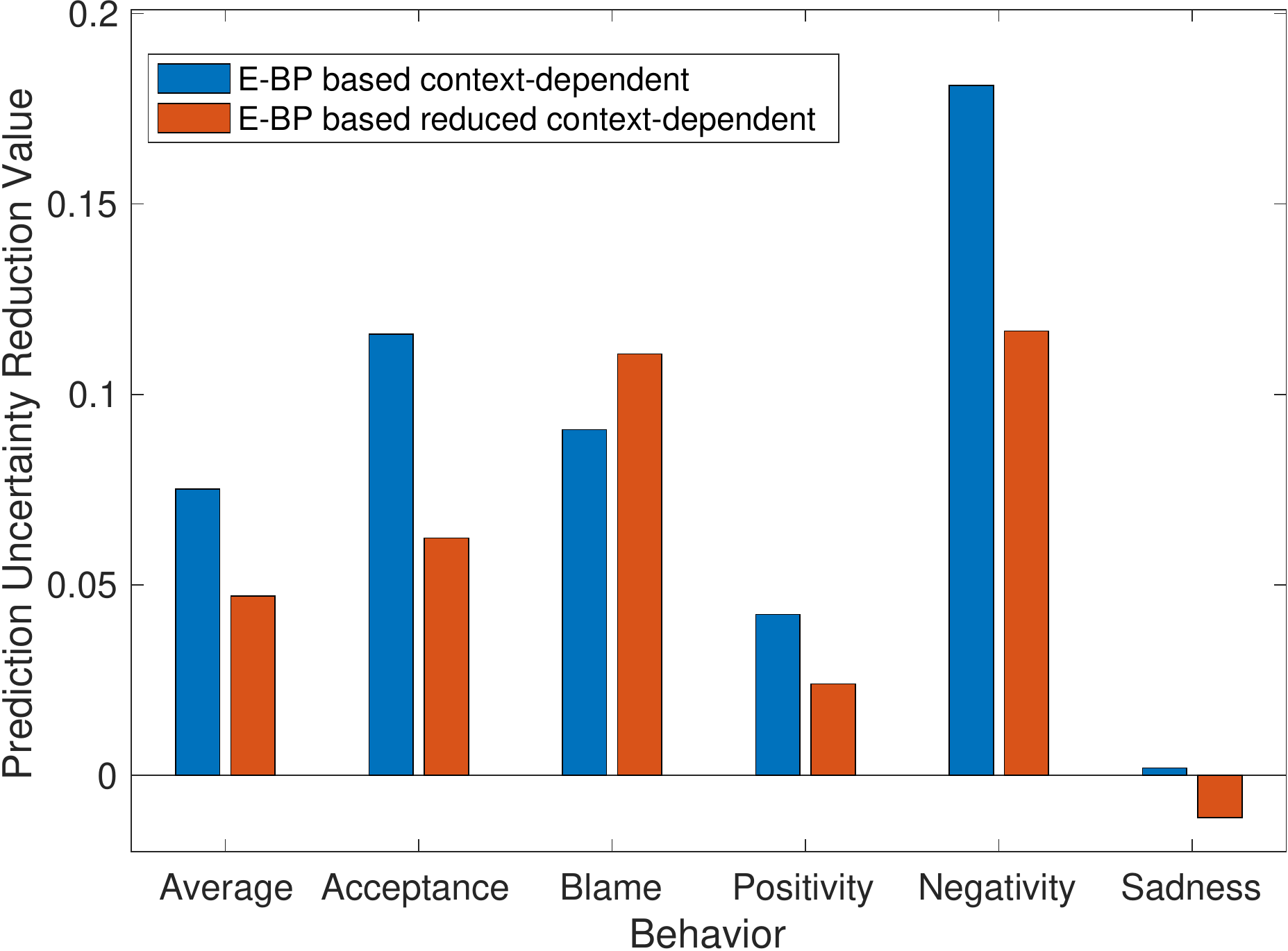}
    \caption{PUR optimal value of \erbpS based context-dependent and reduced context-dependent models across behaviors}
    \label{fig:figure8}
\end{figure}

In Figure \ref{fig:figure8}, as expected, for most behaviors the positive PUR values verify the improvement from using informative \erbpS to simple binary \ecbpS. 
In addition, the results support the hypothesis that the sequential order of affective states is one non-negligible factor of behavior analysis since the PRU of context-dependent (blue color) model is better than that of reduced context one (red color) for most behaviors.

More interestingly, for each behavior, the difference between two bars (\ie PUR difference) can imply the necessity and importance of the sequential and contextual factor of quantifying that behavior. 
We notice that for ``positive'' or more ``complex problem solving'' related behaviors (\eg Acceptance, Positivity), the context based model can achieve better performance than the reduced context model. 
While the PUR differences from ``negative'' related behaviors (\eg Blame, Negativity) varies from different behaviors. 
For example, the behavior of acceptance, with a large PUR difference, it is more related to ``understanding, respect for partner’s views, feelings and behaviors'', which could involve more turns in a dialog and context information. In addition, positivity requires the monitoring of consistent positive behavior, since a single negative instance within a long positive time interval would still reduce positivity to a very low rating. 

In contrast, we see that although blame can still benefit from a larger contextual window, there is no benefit to employing the full context. This may infer that blame expression is more localized. 

Furthermore, our findings are also congruent with many domain annotation processes: Some behaviors are potentially dominated by salient information with short range, and one short duration appearance can have a significant impact on the whole behavior rating, while some behaviors need longer context to analyze \citep{heavey2002couples,jones1998couples}. 

However, among all behaviors, ``sadness'' is always the hardest one to predict with high accuracy, and there is little improvement after introducing different BPs. 
This could be resulting from the extremely skewed distribution towards low ratings as mentioned in above and \citep{georgiou2011aggravating, black2013toward}, which leads to a very blurred binary classification boundary compared to other behaviors.

\section*{Conclusion and Future Work} \label{conclusion}
In this work, we explored the relationship between emotion and behavior states, and further employed emotions as behavioral primitives in behavior classification. 
In our designed systems, we first verified the existing connection between basic emotions and behaviors, then further verified the effectiveness of utilizing emotions as behavior primitive embeddings for behavior quantification through transfer learning. 
Moreover, we designed a reduced context model to investigate the importance of context information in behavior quantification.

Through our models, we additionally investigated the empirical analysis window size for speech behavior understanding, and verified the hypothesis that the order of affective states is an important factor for behavior analysis.
We provided experimental evidence and systematic analyses for behavior understanding via emotion information. 

To summarized, we investigated three questions and we concluded:
\begin{enumerate}
  \item Can the basic emotion states infer behaviors?\\
  The answer is yes. Behavioral states can be weakly inferred from emotions states. However behavior requires richer information than just binary emotions.
  
  \item Can emotion-informed embeddings be employed in the prediction of behaviors?\\ 
  The answer is yes. The rich emotion involved embedding representation helps the  prediction of behaviors. They also do so much better than the information-bottlenecked binary emotions.
  
  \item Is the contextual (sequential) information important in defining behaviors? \\
    The answer is yes. We verify the importance of context of behavior indicators for all behaviors. Some behaviors benefit from incorporating the full interaction (10 minutes) length while others require as little as 16 seconds of information, but all perform best when given contextual information.
\end{enumerate}
Moreover, the proposed neural network systems are not limited to the datasets and domains of this work, but potentially provides a path for investigating a range of problems, such as local versus global, sequential versus non-sequential comparisons in many related areas. 
In addition to the relationship of emotions to behaviors, a range of other cues can also be incorporated towards behavior quantification.  Moreover, many other aspects of behavior, such as entrainment, turn-taking duration, pauses, non-verbal vocalizations, and influence between interlocutors, can be incorporated. 
Many such additional features can be similarly developed on different data and employed as primitives;  for example entrainment measures can be trained through unlabeled data \citep{nasir2018_towards-an-unsu}.

Furthermore, we expect that the results of behavior classification accuracy maybe be further improved through improved architectures, parameter tuning, and data engineering for each behavior of interest. 
In addition, behavior primitives, \eg from emotions, can also be employed via the lexical and visual modalities. 


\bibliography{sample}

\begin{thebibliography}{}

\bibitem[Aldeneh and Provost, 2017]{aldeneh2017using}
Aldeneh, Z. and Provost, E.~M. (2017).
\newblock Using regional saliency for speech emotion recognition.
\newblock In {\em 2017 IEEE International Conference on Acoustics, Speech and
  Signal Processing (ICASSP)}, pages 2741--2745. IEEE.

\bibitem[Ambady and Rosenthal, 1992]{ambady1992thin}
Ambady, N. and Rosenthal, R. (1992).
\newblock Thin slices of expressive behavior as predictors of interpersonal
  consequences: A meta-analysis.
\newblock {\em Psychological bulletin}, 111(2):256.

\bibitem[Anand and Verma, 2015]{anand2015convoluted}
Anand, N. and Verma, P. (2015).
\newblock Convoluted feelings convolutional and recurrent nets for detecting
  emotion from audio data.
\newblock In {\em Technical Report}. Stanford University.

\bibitem[Baer et~al., 2009]{baer2009agency}
Baer, J.~S., Wells, E.~A., Rosengren, D.~B., Hartzler, B., Beadnell, B., and
  Dunn, C. (2009).
\newblock Agency context and tailored training in technology transfer: A pilot
  evaluation of motivational interviewing training for community counselors.
\newblock {\em Journal of substance abuse treatment}, 37(2):191--202.

\bibitem[Baumeister et~al., 2010]{baumeister2010does}
Baumeister, R.~F., DeWall, C.~N., Vohs, K.~D., and Alquist, J.~L. (2010).
\newblock Does emotion cause behavior (apart from making people do stupid,
  destructive things).
\newblock {\em Then a miracle occurs: Focusing on behavior in social
  psychological theory and research}, pages 12--27.

\bibitem[Baumeister et~al., 2007]{baumeister2007emotion}
Baumeister, R.~F., Vohs, K.~D., Nathan~DeWall, C., and Zhang, L. (2007).
\newblock How emotion shapes behavior: Feedback, anticipation, and reflection,
  rather than direct causation.
\newblock {\em Personality and social psychology review}, 11(2):167--203.

\bibitem[Beale and Peter, 2008]{beale2008affect}
Beale, R. and Peter, C. (2008).
\newblock {\em Affect and emotion in human-computer interaction}.
\newblock Springer.

\bibitem[Bengio, 2012]{bengio2012deep}
Bengio, Y. (2012).
\newblock Deep learning of representations for unsupervised and transfer
  learning.
\newblock In {\em Proceedings of ICML Workshop on Unsupervised and Transfer
  Learning}, pages 17--36.

\bibitem[Bengio et~al., 2013]{bengio2013representation}
Bengio, Y., Courville, A., and Vincent, P. (2013).
\newblock Representation learning: A review and new perspectives.
\newblock {\em IEEE transactions on pattern analysis and machine intelligence},
  35(8):1798--1828.

\bibitem[Black et~al., 2010]{black2010automatic}
Black, M., Katsamanis, A., Lee, C.-C., Lammert, A.~C., Baucom, B.~R.,
  Christensen, A., Georgiou, P.~G., and Narayanan, S.~S. (2010).
\newblock Automatic classification of married couples' behavior using audio
  features.
\newblock In {\em Eleventh Annual Conference of the International Speech
  Communication Association}.

\bibitem[Black et~al., 2013]{black2013toward}
Black, M.~P., Katsamanis, A., Baucom, B.~R., Lee, C.-C., Lammert, A.~C.,
  Christensen, A., Georgiou, P.~G., and Narayanan, S.~S. (2013).
\newblock Toward automating a human behavioral coding system for married
  couples’ interactions using speech acoustic features.
\newblock {\em Speech communication}, 55(1):1--21.

\bibitem[Burum and Goldfried, 2007]{burum2007centrality}
Burum, B.~A. and Goldfried, M.~R. (2007).
\newblock The centrality of emotion to psychological change.
\newblock {\em Clinical Psychology: Science and Practice}, 14(4):407--413.

\bibitem[Busso and Narayanan, 2008]{busso2008expression}
Busso, C. and Narayanan, S.~S. (2008).
\newblock The expression and perception of emotions: Comparing assessments of
  self versus others.
\newblock In {\em Ninth Annual Conference of the International Speech
  Communication Association}.

\bibitem[Cabanac, 2002]{cabanac2002emotion}
Cabanac, M. (2002).
\newblock What is emotion?
\newblock {\em Behavioural processes}, 60(2):69--83.

\bibitem[Carney et~al., 2007]{carney2007thin}
Carney, D.~R., Colvin, C.~R., and Hall, J.~A. (2007).
\newblock A thin slice perspective on the accuracy of first impressions.
\newblock {\em Journal of Research in Personality}, 41(5):1054--1072.

\bibitem[Carrillo et~al., 2016]{carrillo2016emotional}
Carrillo, F., Mota, N., Copelli, M., Ribeiro, S., Sigman, M., Cecchi, G., and
  Slezak, D.~F. (2016).
\newblock Emotional intensity analysis in bipolar subjects.
\newblock {\em arXiv preprint arXiv:1606.02231}.

\bibitem[Chakravarthula et~al., 2019]{sandeepJournal2019}
Chakravarthula, S.~N., Baucom, B., Narayanan, S., and Georgiou, P. (2019).
\newblock An analysis of conversational language observation requirements for
  automatic perception of human behaviors in dyadic interactions.
\newblock {\em arXiv}.

\bibitem[Christensen et~al., 2004]{christensen2004traditional}
Christensen, A., Atkins, D.~C., Berns, S., Wheeler, J., Baucom, D.~H., and
  Simpson, L.~E. (2004).
\newblock Traditional versus integrative behavioral couple therapy for
  significantly and chronically distressed married couples.
\newblock {\em Journal of consulting and clinical psychology}, 72(2):176.

\bibitem[Chung et~al., 2014]{chung2014empirical}
Chung, J., Gulcehre, C., Cho, K., and Bengio, Y. (2014).
\newblock Empirical evaluation of gated recurrent neural networks on sequence
  modeling.
\newblock In {\em NIPS 2014 Workshop on Deep Learning, December 2014}.

\bibitem[Collobert et~al., 2011]{collobert2011natural}
Collobert, R., Weston, J., Bottou, L., Karlen, M., Kavukcuoglu, K., and Kuksa,
  P. (2011).
\newblock Natural language processing (almost) from scratch.
\newblock {\em Journal of machine learning research}, 12(Aug):2493--2537.

\bibitem[Cowie and Cornelius, 2003]{cowie2003describing}
Cowie, R. and Cornelius, R.~R. (2003).
\newblock Describing the emotional states that are expressed in speech.
\newblock {\em Speech communication}, 40(1-2):5--32.

\bibitem[Cowie et~al., 2001]{cowie2001emotion}
Cowie, R., Douglas-Cowie, E., Tsapatsoulis, N., Votsis, G., Kollias, S.,
  Fellenz, W., and Taylor, J.~G. (2001).
\newblock Emotion recognition in human-computer interaction.
\newblock {\em IEEE Signal processing magazine}, 18(1):32--80.

\bibitem[Cummins et~al., 2015]{cummins2015review}
Cummins, N., Scherer, S., Krajewski, J., Schnieder, S., Epps, J., and Quatieri,
  T.~F. (2015).
\newblock A review of depression and suicide risk assessment using speech
  analysis.
\newblock {\em Speech Communication}, 71:10--49.

\bibitem[Dunlop et~al., 2008]{dunlop2008can}
Dunlop, S., Wakefield, M., and Kashima, Y. (2008).
\newblock Can you feel it? negative emotion, risk, and narrative in health
  communication.
\newblock {\em Media Psychology}, 11(1):52--75.

\bibitem[Ekman, 1992a]{ekman1992_are-there-basic}
Ekman, P. (1992a).
\newblock Are there basic emotions?

\bibitem[Ekman, 1992b]{ekman1992_an-argument-for}
Ekman, P. (1992b).
\newblock An argument for basic emotions.
\newblock {\em Cognition \& emotion}, 6(3-4):169--200.

\bibitem[El~Ayadi et~al., 2011]{el2011survey}
El~Ayadi, M., Kamel, M.~S., and Karray, F. (2011).
\newblock Survey on speech emotion recognition: Features, classification
  schemes, and databases.
\newblock {\em Pattern Recognition}, 44(3):572--587.

\bibitem[Feinberg et~al., 2007]{feinberg2007longitudinal}
Feinberg, M.~E., Kan, M.~L., and Hetherington, E.~M. (2007).
\newblock The longitudinal influence of coparenting conflict on parental
  negativity and adolescent maladjustment.
\newblock {\em Journal of Marriage and Family}, 69(3):687--702.

\bibitem[Georgiou et~al., 2011a]{georgiou2011aggravating}
Georgiou, P.~G., Black, M.~P., Lammert, A.~C., Baucom, B.~R., and Narayanan,
  S.~S. (2011a).
\newblock ``that's aggravating, very aggravating'': Is it possible to classify
  behaviors in couple interactions using automatically derived lexical
  features?
\newblock In D'Mello, S., Graesser, A., Schuller, B., and Martin, J.-C.,
  editors, {\em Affective Computing and Intelligent Interaction}, pages 87--96,
  Berlin, Heidelberg. Springer Berlin Heidelberg.

\bibitem[Georgiou et~al., 2011b]{georgiou2011behavioral}
Georgiou, P.~G., Black, M.~P., and Narayanan, S.~S. (2011b).
\newblock Behavioral signal processing for understanding (distressed) dyadic
  interactions: some recent developments.
\newblock In {\em Proceedings of the 2011 joint ACM workshop on Human gesture
  and behavior understanding}, pages 7--12. ACM.

\bibitem[Ghahremani et~al., 2014]{ghahremani2014pitch}
Ghahremani, P., BabaAli, B., Povey, D., Riedhammer, K., Trmal, J., and
  Khudanpur, S. (2014).
\newblock A pitch extraction algorithm tuned for automatic speech recognition.
\newblock In {\em Acoustics, Speech and Signal Processing (ICASSP), 2014 IEEE
  International Conference on}, pages 2494--2498. IEEE.

\bibitem[Ghosh et~al., 2011]{ghosh2011robust}
Ghosh, P.~K., Tsiartas, A., and Narayanan, S. (2011).
\newblock Robust voice activity detection using long-term signal variability.
\newblock {\em IEEE Transactions on Audio, Speech, and Language Processing},
  19(3):600--613.

\bibitem[Gupta et~al., 2014]{gupta2014multimodal}
Gupta, R., Malandrakis, N., Xiao, B., Guha, T., Van~Segbroeck, M., Black, M.,
  Potamianos, A., and Narayanan, S. (2014).
\newblock Multimodal prediction of affective dimensions and depression in
  human-computer interactions.
\newblock In {\em Proceedings of the 4th International Workshop on Audio/Visual
  Emotion Challenge}, pages 33--40. ACM.

\bibitem[Han et~al., 2014]{han2014speech}
Han, K., Yu, D., and Tashev, I. (2014).
\newblock Speech emotion recognition using deep neural network and extreme
  learning machine.
\newblock In {\em Fifteenth annual conference of the international speech
  communication association}.

\bibitem[Heavey et~al., 2002]{heavey2002couples}
Heavey, C., Gill, D., and Christensen, A. (2002).
\newblock Couples interaction rating system 2 (cirs2).
\newblock {\em University of California, Los Angeles}, 7.

\bibitem[Heavey et~al., 1995]{heavey1995longitudinal}
Heavey, C.~L., Christensen, A., and Malamuth, N.~M. (1995).
\newblock The longitudinal impact of demand and withdrawal during marital
  conflict.
\newblock {\em Journal of consulting and clinical psychology}, 63(5):797.

\bibitem[Heyman, 2004]{heyman2004rapid}
Heyman, R.~E. (2004).
\newblock Rapid marital interaction coding system (rmics).
\newblock In {\em Couple observational coding systems}, pages 81--108.
  Routledge.

\bibitem[Heyman et~al., 2001]{heyman2001much}
Heyman, R.~E., Chaudhry, B.~R., Treboux, D., Crowell, J., Lord, C., Vivian, D.,
  and Waters, E.~B. (2001).
\newblock How much observational data is enough? an empirical test using
  marital interaction coding.
\newblock {\em Behavior therapy}, 32(1):107--122.

\bibitem[Hoff, 2009]{hoff2009language}
Hoff, E. (2009).
\newblock Language development at an early age: Learning mechanisms and
  outcomes from birth to five years.
\newblock {\em Encyclopedia on early childhood development}, pages 1--5.

\bibitem[Huang et~al., 2017]{huang2017characterizing}
Huang, C.-W., Narayanan, S., et~al. (2017).
\newblock Characterizing types of convolution in deep convolutional recurrent
  neural networks for robust speech emotion recognition.
\newblock {\em arXiv preprint arXiv:1706.02901}.

\bibitem[Huang and Narayanan, 2017]{huang2017deep}
Huang, C.-W. and Narayanan, S.~S. (2017).
\newblock Deep convolutional recurrent neural network with attention mechanism
  for robust speech emotion recognition.
\newblock In {\em Multimedia and Expo (ICME), 2017 IEEE International
  Conference on}, pages 583--588. IEEE.

\bibitem[Jones and Christensen, 1998]{jones1998couples}
Jones, J. and Christensen, A. (1998).
\newblock Couples interaction study: Social support interaction rating system.
\newblock {\em University of California, Los Angeles}, 7.

\bibitem[Katsamanis et~al., 2011]{katsamanis2011sailalign}
Katsamanis, A., Black, M., Georgiou, P.~G., Goldstein, L., and Narayanan, S.
  (2011).
\newblock Sailalign: Robust long speech-text alignment.
\newblock In {\em Proc. of Workshop on New Tools and Methods for Very-Large
  Scale Phonetics Research}.

\bibitem[Khorram et~al., 2018]{khorram2018priori}
Khorram, S., Jaiswal, M., Gideon, J., McInnis, M., and Provost, E.-M. (2018).
\newblock The priori emotion dataset: Linking mood to emotion detected
  in-the-wild.
\newblock {\em Proc. Interspeech 2018}, pages 1903--1907.

\bibitem[Kingma and Ba, 2014]{kingma2014adam}
Kingma, D.~P. and Ba, J. (2014).
\newblock Adam: A method for stochastic optimization.
\newblock {\em arXiv preprint arXiv:1412.6980}.

\bibitem[Le and Provost, 2013]{le2013emotion}
Le, D. and Provost, E.~M. (2013).
\newblock Emotion recognition from spontaneous speech using hidden markov
  models with deep belief networks.
\newblock In {\em Automatic Speech Recognition and Understanding (ASRU), 2013
  IEEE Workshop on}, pages 216--221. IEEE.

\bibitem[Lee and Tashev, 2015]{lee2015high}
Lee, J. and Tashev, I. (2015).
\newblock High-level feature representation using recurrent neural network for
  speech emotion recognition.
\newblock In {\em Sixteenth Annual Conference of the International Speech
  Communication Association}.

\bibitem[Li et~al., 2016]{li2016_sparsely-connec}
Li, H., Baucom, B., and Georgiou, P. (2016).
\newblock Sparsely connected and disjointly trained deep neural networks for
  low resource behavioral annotation: Acoustic classification in couples'
  therapy.
\newblock In {\em Proceedings of Interspeech}, San Francisco, CA.

\bibitem[Li et~al., 2017]{li2017unsupervised}
Li, H., Baucom, B., and Georgiou, P. (2017).
\newblock Unsupervised latent behavior manifold learning from acoustic
  features: Audio2behavior.
\newblock In {\em 2017 IEEE International Conference on Acoustics, Speech and
  Signal Processing (ICASSP)}, pages 5620--5624. IEEE.

\bibitem[Lim et~al., 2016]{lim2016speech}
Lim, W., Jang, D., and Lee, T. (2016).
\newblock Speech emotion recognition using convolutional and recurrent neural
  networks.
\newblock In {\em Signal and information processing association annual summit
  and conference (APSIPA), 2016 Asia-Pacific}, pages 1--4. IEEE.

\bibitem[Lustgarten, 2015]{lustgarten2015emerging}
Lustgarten, S.~D. (2015).
\newblock Emerging ethical threats to client privacy in cloud communication and
  data storage.
\newblock {\em Professional Psychology: Research and Practice}, 46(3):154.

\bibitem[Mao et~al., 2014]{mao2014learning}
Mao, Q., Dong, M., Huang, Z., and Zhan, Y. (2014).
\newblock Learning salient features for speech emotion recognition using
  convolutional neural networks.
\newblock {\em IEEE Transactions on Multimedia}, 16(8):2203--2213.

\bibitem[Metallinou et~al., 2012]{metallinou2012context}
Metallinou, A., Wollmer, M., Katsamanis, A., Eyben, F., Schuller, B., and
  Narayanan, S. (2012).
\newblock Context-sensitive learning for enhanced audiovisual emotion
  classification.
\newblock {\em IEEE Transactions on Affective Computing}, 3(2):184--198.

\bibitem[Mower and Narayanan, 2011]{mower2011hierarchical}
Mower, E. and Narayanan, S. (2011).
\newblock A hierarchical static-dynamic framework for emotion classification.
\newblock In {\em Acoustics, Speech and Signal Processing (ICASSP), 2011 IEEE
  International Conference on}, pages 2372--2375. IEEE.

\bibitem[Narayanan and Georgiou, 2013]{narayanan2013behavioral}
Narayanan, S. and Georgiou, P.~G. (2013).
\newblock Behavioral signal processing: Deriving human behavioral informatics
  from speech and language.
\newblock {\em Proceedings of the IEEE}, 101(5):1203--1233.

\bibitem[Nasir et~al., 2018]{nasir2018_towards-an-unsu}
Nasir, M., Baucom, B., Narayanan, S., and Georgiou, P. (2018).
\newblock Towards an unsupervised entrainment distance in conversational speech
  using deep neural networks.
\newblock In {\em Interspeech / arXiv:1804.08782}.

\bibitem[Nasir et~al., 2017a]{nasir2017_complexity-in-s}
Nasir, M., Baucom, B.~R., Bryan, C.~J., Narayanan, S., and Georgiou, P.
  (2017a).
\newblock Complexity in speech and its relation to emotional bond in
  therapist-patient interactions during suicide risk assessment interviews.
\newblock In {\em Interspeech}, Stockholm, Sweden.

\bibitem[Nasir et~al., 2017b]{nasir2017predicting}
Nasir, M., Baucom, B.~R., Georgiou, P., and Narayanan, S. (2017b).
\newblock Predicting couple therapy outcomes based on speech acoustic features.
\newblock {\em PloS one}, 12(9):e0185123.

\bibitem[Nasir et~al., 2016]{nasir2016multimodal}
Nasir, M., Jati, A., Shivakumar, P.~G., Nallan~Chakravarthula, S., and
  Georgiou, P. (2016).
\newblock Multimodal and multiresolution depression detection from speech and
  facial landmark features.
\newblock In {\em Proceedings of the 6th International Workshop on Audio/Visual
  Emotion Challenge}, pages 43--50. ACM.

\bibitem[Oatley and Jenkins, 1996]{oatley1996understanding}
Oatley, K. and Jenkins, J.~M. (1996).
\newblock {\em Understanding emotions.}
\newblock Blackwell publishing.

\bibitem[Picard, 2003]{picard2003affective}
Picard, R.~W. (2003).
\newblock Affective computing: challenges.
\newblock {\em International Journal of Human-Computer Studies},
  59(1-2):55--64.

\bibitem[Povey et~al., 2011]{povey2011kaldi}
Povey, D., Ghoshal, A., Boulianne, G., Burget, L., Glembek, O., Goel, N.,
  Hannemann, M., Motlicek, P., Qian, Y., Schwarz, P., et~al. (2011).
\newblock The kaldi speech recognition toolkit.
\newblock In {\em IEEE 2011 workshop on automatic speech recognition and
  understanding}, number EPFL-CONF-192584. IEEE Signal Processing Society.

\bibitem[Sander and Scherer, 2014]{sander2014oxford}
Sander, D. and Scherer, K. (2014).
\newblock {\em Oxford companion to emotion and the affective sciences}.
\newblock OUP Oxford.

\bibitem[Schacter et~al., 2011]{WorthPsychology}
Schacter, D., Gilbert, D.~T., and Wegner, D.~M. (2011).
\newblock {\em Psychology (2nd Edition)}.
\newblock Worth, New York.

\bibitem[Scherer, 2005]{scherer2005emotions}
Scherer, K.~R. (2005).
\newblock What are emotions? and how can they be measured?
\newblock {\em Social science information}, 44(4):695--729.

\bibitem[Schlosberg, 1954]{schlosberg1954three}
Schlosberg, H. (1954).
\newblock Three dimensions of emotion.
\newblock {\em Psychological review}, 61(2):81.

\bibitem[Schuller et~al., 2011]{schuller2011recognising}
Schuller, B., Batliner, A., Steidl, S., and Seppi, D. (2011).
\newblock Recognising realistic emotions and affect in speech: State of the art
  and lessons learnt from the first challenge.
\newblock {\em Speech Communication}, 53(9-10):1062--1087.

\bibitem[Schuller et~al., 2003]{schuller2003hidden}
Schuller, B., Rigoll, G., and Lang, M. (2003).
\newblock Hidden markov model-based speech emotion recognition.
\newblock In {\em Acoustics, Speech, and Signal Processing, 2003.
  Proceedings.(ICASSP'03). 2003 IEEE International Conference on}, volume~2,
  pages II--1. IEEE.

\bibitem[Schuller, 2018]{schuller2018speech}
Schuller, B.~W. (2018).
\newblock Speech emotion recognition: two decades in a nutshell, benchmarks,
  and ongoing trends.
\newblock {\em Communications of the ACM}, 61(5):90--99.

\bibitem[Sculley et~al., 2015]{sculley2015hidden}
Sculley, D., Holt, G., Golovin, D., Davydov, E., Phillips, T., Ebner, D.,
  Chaudhary, V., Young, M., Crespo, J.-F., and Dennison, D. (2015).
\newblock Hidden technical debt in machine learning systems.
\newblock In {\em Advances in neural information processing systems}, pages
  2503--2511.

\bibitem[Soken and Pick, 1999]{soken1999infants}
Soken, N.~H. and Pick, A.~D. (1999).
\newblock Infants' perception of dynamic affective expressions: Do infants
  distinguish specific expressions?
\newblock {\em Child development}, 70(6):1275--1282.

\bibitem[Soltau et~al., 2017]{soltau2017neural}
Soltau, H., Liao, H., and Sak, H. (2017).
\newblock Neural speech recognizer: Acoustic-to-word lstm model for large
  vocabulary speech recognition.
\newblock {\em Proc. Interspeech 2017}, pages 3707--3711.

\bibitem[Spector and Fox, 2002]{spector2002emotion}
Spector, P.~E. and Fox, S. (2002).
\newblock An emotion-centered model of voluntary work behavior: Some parallels
  between counterproductive work behavior and organizational citizenship
  behavior.
\newblock {\em Human resource management review}, 12(2):269--292.

\bibitem[Srivastava et~al., 2014]{srivastava2014dropout}
Srivastava, N., Hinton, G., Krizhevsky, A., Sutskever, I., and Salakhutdinov,
  R. (2014).
\newblock Dropout: a simple way to prevent neural networks from overfitting.
\newblock {\em The Journal of Machine Learning Research}, 15(1):1929--1958.

\bibitem[Stasak et~al., 2016]{stasak2016investigation}
Stasak, B., Epps, J., Cummins, N., and Goecke, R. (2016).
\newblock An investigation of emotional speech in depression classification.
\newblock In {\em Interspeech}, pages 485--489.

\bibitem[Tanaka et~al., 2017]{tanaka2017brain}
Tanaka, T., Yamamoto, T., and Haruno, M. (2017).
\newblock Brain response patterns to economic inequity predict present and
  future depression indices.
\newblock {\em Nature Human Behaviour}, 1(10):748.

\bibitem[Tao and Tan, 2005]{tao2005affective}
Tao, J. and Tan, T. (2005).
\newblock Affective computing: A review.
\newblock In {\em International Conference on Affective computing and
  intelligent interaction}, pages 981--995. Springer.

\bibitem[Tong et~al., 2017]{tong2017combating}
Tong, E., Zadeh, A., Jones, C., and Morency, L.-P. (2017).
\newblock Combating human trafficking with multimodal deep models.
\newblock In {\em Proceedings of the 55th Annual Meeting of the Association for
  Computational Linguistics (Volume 1: Long Papers)}, volume~1, pages
  1547--1556.

\bibitem[Torrey and Shavlik, 2010]{torrey2010transfer}
Torrey, L. and Shavlik, J. (2010).
\newblock Transfer learning.
\newblock In {\em Handbook of Research on Machine Learning Applications and
  Trends: Algorithms, Methods, and Techniques}, pages 242--264. IGI Global.

\bibitem[Tseng et~al., 2018]{tseng2018multi}
Tseng, S.-Y., Baucom, B., and Georgiou, P. (2018).
\newblock Unsupervised online multitask learning of behavioral sentence
  embeddings.
\newblock {\em arXiv preprint arXiv:1807.06792}.

\bibitem[Tseng et~al., 2016]{tseng2016_couples-behavio}
Tseng, S.-Y., Chakravarthula, S.~N., Baucom, B., and Georgiou, P. (2016).
\newblock Couples behavior modeling and annotation using low-resource {LSTM}
  language models.
\newblock In {\em Proceedings of Interspeech}, San Francisco, CA.

\bibitem[Venek et~al., 2017]{venek2017adolescent}
Venek, V., Scherer, S., Morency, L.-P., Pestian, J., et~al. (2017).
\newblock Adolescent suicidal risk assessment in clinician-patient interaction.
\newblock {\em IEEE Transactions on Affective Computing}, 8(2):204--215.

\bibitem[Vinciarelli et~al., 2009]{vinciarelli2009social}
Vinciarelli, A., Pantic, M., and Bourlard, H. (2009).
\newblock Social signal processing: Survey of an emerging domain.
\newblock {\em Image and vision computing}, 27(12):1743--1759.

\bibitem[W{\"o}llmer et~al., 2010]{wollmer2010context}
W{\"o}llmer, M., Metallinou, A., Eyben, F., Schuller, B., and Narayanan, S.
  (2010).
\newblock Context-sensitive multimodal emotion recognition from speech and
  facial expression using bidirectional lstm modeling.
\newblock In {\em Proc. INTERSPEECH 2010, Makuhari, Japan}, pages 2362--2365.

\bibitem[Zadeh, 2019]{a2zadeh_2019}
Zadeh, A.~B. (2019).
\newblock Cmu-multimodalsdk.
\newblock Available at https://github.com/A2Zadeh/CMU-MultimodalSDK (accessed
  March 2019).

\bibitem[Zadeh et~al., 2018]{zadeh2018multimodal}
Zadeh, A.~B., Liang, P.~P., Poria, S., Cambria, E., and Morency, L.-P. (2018).
\newblock Multimodal language analysis in the wild: Cmu-mosei dataset and
  interpretable dynamic fusion graph.
\newblock In {\em Proceedings of the 56th Annual Meeting of the Association for
  Computational Linguistics (Volume 1: Long Papers)}, volume~1, pages
  2236--2246.

\bibitem[Zheng et~al., 2015]{zheng2015experimental}
Zheng, W., Yu, J., and Zou, Y. (2015).
\newblock An experimental study of speech emotion recognition based on deep
  convolutional neural networks.
\newblock In {\em Affective Computing and Intelligent Interaction (ACII), 2015
  International Conference on}, pages 827--831. IEEE.

\end{thebibliography}

\newpage
\appendix

\section{Detailed network architecture and training parameters}
\begin{table}[htb]
\centering
\begin{tabular}{l}
\hline
Multi-Emotion Regression Network (ER) framework (Input: 84 * 100 ; Output:6) \\ \hline
Training details: Adam optimizer(lr = 1e-05), batch size 16, MSELoss                  \\ \hline
Conv1d(in\_ch=84, out\_ch=96, kernel size=10, stride=2, padding=0) ReLU                \\ 
Conv1d(in\_ch=96, out\_ch=96, kernel size=5, stride=2, padding=0)  ReLU                \\ 
Conv1d(in\_ch=96, out\_ch=96, kernel size=5, stride=2, padding=0)  ReLU                \\ 
Conv1d(in\_ch=96, out\_ch=128, kernel size=3, stride=2, padding=0) ReLU                \\ 
AdaptiveMaxPool1d(1)                                                                 \\ 
Linear(in =128, out =128) ReLU                                                       \\ 
Linear(in =128, out =128) ReLU                                                       \\ 
Linear(in =128, out =6)                                                              \\ \hline
\end{tabular}
\caption{Network architecture of ER}
\label{tab:NN_API}
\end{table}

\begin{table}[htb]
\centering
\begin{tabular}{l|c}
\hline
Single-Emotion Classification Network (EC) framework (Input: 84 * 100 ; Output: 2)                              &                             \\ \cline{1-1}
Training details: Adam optimizer(lr = 1e-05),  CrossEntropyLoss, batch size: 32; 64; 128 &                             \\ \hline
Conv1d(in\_ch=84, out\_ch=96, kernel size=10, stride=2, padding=0) ReLU                    & \multirow{7}{*}{Pretrained} \\ 
Conv1d(in\_ch=96, out\_ch=96, kernel size=5, stride=2, padding=0)  ReLU                    &                             \\ 
Conv1d(in\_ch=96, out\_ch=96, kernel size=5, stride=2, padding=0)  ReLU                    &                             \\ 
Conv1d(in\_ch=96, out\_ch=128, kernel size=3, stride=2, padding=0) ReLU                    &                             \\ 
AdaptiveMaxPool1d(1)                                                                     &                             \\ 
Linear(in =128, out =128) ReLU                                                           &                             \\ 
Linear(in =128, out =128) ReLU                                                           &                             \\ \hline
Linear(in =128, out =64) PReLU                                                           & \multirow{3}{*}{Trainable}  \\ 
Linear(in =64, out =64) PReLU                                                            &                             \\ 
Linear(in =64, out =2)                                                                   &                             \\ \hline
\end{tabular}
\caption{Network architecture of EC}
\label{tab:NN_emo_recognition}
\end{table}

\begin{table}[htb]
\centering
\begin{tabular}{l|c}
\hline
\ecbpS based context-dependent behavior recognition model (Input: seq\_len*6; Output: 5)                                                                                                                  &                                \\ \cline{1-1}
\begin{tabular}[c]{@{}l@{}}Training details: Adam optimizer(lr = 1e-04) + Polynomial learning rate decay,  \\ Masked BCEWithLogitsLoss, batch size: 1\end{tabular} &                                \\ \hline
Emotion recognition framework                                                                                                                                      & \multicolumn{1}{c}{Pretrained} \\ \hline
GRU(in\_size =6, hidden\_size = 128, num\_layers=2)                                                                                                                      & \multirow{3}{*}{Trainable}     \\ 
Linear(in =128, out =64) ReLU                                                                                                                                      &                                \\ 
Linear(in =64, out =5)                                                                                                                                             &                                \\ \hline
\end{tabular}
\caption{\ecbpS based context-dependent behavior recognition model framework}
\label{tab:NN_emo2beh_seq}
\end{table}

\begin{table}[htb]
\centering
\scalebox{0.85}{
\begin{tabular}{l|c}
\hline
\erbpS based context-dependent behavior recognition model (Input: seq\_len * 84 * 100 ; Output: 5)                                                                                                                    & \multirow{2}{*}{}                                                                             \\ \cline{1-1}
\begin{tabular}[c]{@{}l@{}}Training details: Adam optimizer(lr = 1e-05) + Polynomial learning rate decay,  \\ Masked BCEWithLogitsLoss, batch size: 1, epochs=300\end{tabular} &                                                                                               \\ \hline
Conv1d(in\_ch=84, out\_ch=96, kernel size=10, stride=2, padding=0) ReLU                                                                                                          & \multirow{5}{*}{\begin{tabular}[c]{@{}l@{}}Partly pretrained\\ Partly trainable\end{tabular}} \\ 
Conv1d(in\_ch=96, out\_ch=96, kernel size=5, stride=2, padding=0)  ReLU                                                                                                          &                                                                                               \\ 
Conv1d(in\_ch=96, out\_ch=96, kernel size=5, stride=2, padding=0)  ReLU                                                                                                          &                                                                                               \\ 
Conv1d(in\_ch=96, out\_ch=128, kernel size=3, stride=2, padding=0) ReLU                                                                                                          &                                                                                               \\ 
AdaptiveMaxPool1d(1)                                                                                                                                                           &                                                                                               \\ \hline
GRU(in\_size =128, hidden\_size = 128, num\_layers=2)                                                                                                                                  & \multirow{3}{*}{Trainable}                                                                    \\ 
Linear(in =128, out =64) ReLU                                                                                                                                                  &                                                                                               \\
Linear(in =64, out =5)                                                                                                                                                         &                                                                                               \\ \hline
\end{tabular}
}

\caption{\erbpS based context-dependent behavior recognition model framework}
\label{tab:NN_emoebd2beh_seq}
\end{table}

\begin{table}[]
\centering
\scalebox{0.85}{
\begin{tabular}{l|c}
\hline
\begin{tabular}[c]{@{}l@{}}E-BP based reduced context-dependent behavior recognition model\\ (Input: 84 * seq\_len ; Output: 5)\end{tabular}                                                                                                                                                                                                                                                                                                                                                                                                                                                                                                                                                     &                                                                                     \\ \hline
\begin{tabular}[c]{@{}l@{}}Training details: Adam optimizer(lr = 1e-04) + Polynomial learning rate decay,\\ Masked BCEWithLogitsLoss, batch size: 48, epochs=350\end{tabular}                                                                                                                                                                                                                                                                                                                                                                                                                                                                                                                    &                                                                                     \\ \hline
\begin{tabular}[c]{@{}l@{}}Conv1d(in\_ch=84, out\_ch=96, kernel size=10, stride=2, padding=0) ReLU\\ Conv1d(in\_ch=96, out\_ch=96, kernel size=5, stride=2, padding=0) ReLU\\ Conv1d(in\_ch=96, out\_ch=96, kernel size=5, stride=2, padding=0) ReLU\\ Conv1d(in\_ch=96, out\_ch=128, kernel size=3, stride=2, padding=0) ReLU\end{tabular}                                                                                                                                                                                                                                                                                                                                                      & \begin{tabular}[c]{@{}l@{}}Behavior primitive embedding\\ (Pretrained)\end{tabular} \\ \hline
\begin{tabular}[c]{@{}l@{}}Conv1d(in\_ch=128, out\_ch=96, kernel size=3, stride=2, padding=0)\\ \textbf{AvgPool1d(kernel size=2, stride=2)} ReLU\\ Dropout(prob=0.4)\\ Conv1d(in\_ch=96, out\_ch=96, kernel size=3, stride=2, padding=0)\\ \textbf{AvgPool1d(kernel size=2, stride=2)} ReLU\\ Dropout(prob=0.4)\\ Conv1d(in\_ch=96, out\_ch=96, kernel size=3, stride=1, padding=0)\\ \textbf{AvgPool1d(kernel size=2, stride=2)} ReLU\\ Dropout(prob=0.4)\\ Conv1d(in\_ch=96, out\_ch=128, kernel size=3, stride=1, padding=0)\\ \textbf{AvgPool1d(kernel size=2, stride=2)} ReLU\\ Dropout(prob=0.5)\\ AdaptiveMaxPool1d(1)\\ Linear(in =128, out =128) ReLU\\ Linear(in =128, out =64) ReLU\\ Linear(in =64, out =5)\end{tabular} & Trainable                                                                           \\ \hline
\end{tabular}
}
\caption{\erbpS based reduced context-dependent behavior recognition model framework. Those AvgPool1d layers are optional to adjust temporal receptive field size.}
\label{tab:NN_emoebd2beh_nonseq}
\end{table}


\end{document}